\documentclass[pdflatex,sn-mathphys-num]{sn-jnl}

\usepackage{graphicx}%
\usepackage{multirow}%
\usepackage{amsmath,amssymb,amsfonts}%
\usepackage{amsthm}%
\usepackage{mathrsfs}%
\usepackage[title]{appendix}%
\usepackage{xcolor}%
\usepackage{textcomp}%
\usepackage{manyfoot}%
\usepackage{booktabs}%
\usepackage{algorithm}%
\usepackage{algorithmicx}%
\usepackage{algpseudocode}%
\usepackage{listings}%

\usepackage{xspace}
\usepackage{multirow}
\usepackage{bbding}
\usepackage{graphicx}
\usepackage{booktabs}
\usepackage{xcolor}
\usepackage{colortbl}

\usepackage{amsmath}
\usepackage{algorithm}
\usepackage{algpseudocode}
\usepackage{floatflt} 
\usepackage{wrapfig}
\usepackage{caption}%
\usepackage{hyperref}

\newcommand{\eg}{\textit{e.g.}\xspace}
\newcommand{\etc}{\textit{etc.}\xspace}

\theoremstyle{thmstyleone}%
\theoremstyle{thmstyletwo}%

\theoremstyle{thmstylethree}%

\begin{document}

\title[Article Title]{Towards Any-Quality Image Segmentation via Generative and Adaptive Latent Space Enhancement}


\author[1]{\fnm{Guangqian} \sur{Guo}}\email{guogq21@mail.nwpu.edu.cn}
\author[1]{\fnm{Aixi} \sur{Ren}}\email{renaixi@mail.nwpu.edu.cn}
\author[2]{\fnm{Yong} \sur{Guo}}\email{guoyongcs@gmail.com}
\author[3]{\fnm{Xuehui} \sur{Yu}} \email{yuxuehui17@mails.ucas.ac.cn}
\author[1]{\fnm{Jiacheng} \sur{Tian}} \email{tjc@mail.nwpu.edu.cn}
\author[1]{\fnm{Wenli} \sur{Li}} \email{2024264797@mail.nwpu.edu.cn}
\author[1]{\fnm{Chaowei} \sur{Wang}} \email{chaowei\_wang@mail.nwpu.edu.cn}
\author[1]{\fnm{Yaoxing} \sur{Wang}} \email{wangyx24@mail.nwpu.edu.cn}
\author*[1]{\fnm{Shan} \sur{Gao}} \email{gaoshan@nwpu.edu.cn}

\affil[1]{\orgname{Northwestern Polytechnical University}, \orgaddress{\city{Xi'an}, \country{China}}}
\affil[2]{\orgname{Max Planck Institute for Informatics}, \orgaddress{\city{Saarbrücken}, \country{Germany}}}
\affil[3]{\orgname{University of Chinese Academy of Sciences}, \orgaddress{\city{Beijing}, \country{China}}}



\abstract{
Segment Anything Models (SAMs), known for their
exceptional zero-shot segmentation performance, have garnered significant attention in the research community. 
Nevertheless, their performance drops significantly on severely degraded, low-quality images, limiting their effectiveness in real-world scenarios. To address this, we propose \textbf{GleSAM++}, which utilizes \underline{G}enerative \underline{L}atent space \underline{E}nhancement to boost robustness on low-quality images, thus enabling generalization across various image qualities.  Specifically, we adapt the concept of latent diffusion to SAM-based segmentation frameworks and perform the generative diffusion process in the latent space of SAM to reconstruct a high-quality representation, thereby improving segmentation. 
Additionally, to improve compatibility between the pre-trained diffusion model and the segmentation framework, we introduce two techniques, \textit{i.e.}, \textbf{{Feature Distribution Alignment}} (FDA) and \textbf{{Channel Replication and Expansion}} (CRE).
However, the above components lack explicit guidance regarding the degree of degradation. The model is forced to implicitly fit a complex noise distribution that spans conditions from mild noise to severe artifacts, which substantially increases the learning burden and leads to suboptimal reconstructions.
To address this issue, we further introduce a \textbf{{Degradation-aware Adaptive Enhancement}} (DAE) mechanism. The key principle of DAE is to decouple the reconstruction process for arbitrary-quality features into two stages: degradation-level prediction and degradation-aware reconstruction. This design reduces the optimization difficulty of the model and consequently enhances the effectiveness of feature reconstruction.
Our method can be applied to pre-trained SAM and SAM2 with only minimal additional learnable parameters, allowing for efficient optimization.
We also construct the LQSeg dataset with a greater diversity of degradation types and levels for training and evaluating the model. Extensive experiments demonstrate that GleSAM++ significantly improves segmentation robustness on complex degradations while maintaining generalization to clear images. Furthermore, GleSAM++ also performs well on unseen degradations, underscoring the versatility of our approach and dataset. Codes, datasets, and trained models will be available at the Project Page:  \url{https://guangqian-guo.github.io/glesam++}.
}



\maketitle

\section{Introduction}
\label{sec:intro}

Accurate visual perception ~\cite{maskrcnn, pcod, hybridnet, videoseg, hanet, effective, glesam} that can operate under diverse and unpredictable real-world conditions remains a fundamental task in computer vision research. 
Robust performance is especially critical in various high-level visual applications, such as robotics, autonomous driving, and medical image analysis.
The recently developed Segment Anything Models (SAMs), including SAM \cite{sam} and SAM2 \cite{sam2}, serving as a foundational segmentation model, have gained significant influence within the community ~\cite{rsprompter, sam-med, sam-med2, med-sam2, sa2va} due to their outstanding zero-shot segmentation abilities. 
It can interactively segment any object in an image using visual prompts such as points and bounding boxes.
SAM's robust generalizability has led to breakthroughs and new paradigms in various downstream tasks, including remote sensing~\cite{rsprompter, sam-rs2, sam-rs3}, automatic data annotation~\cite{samrs, sam-label, p2p}, and medical image segmentation~\cite{sam-med, sam-med2}.

\begin{figure}[h]
  \centering
   \includegraphics[width=1\linewidth]{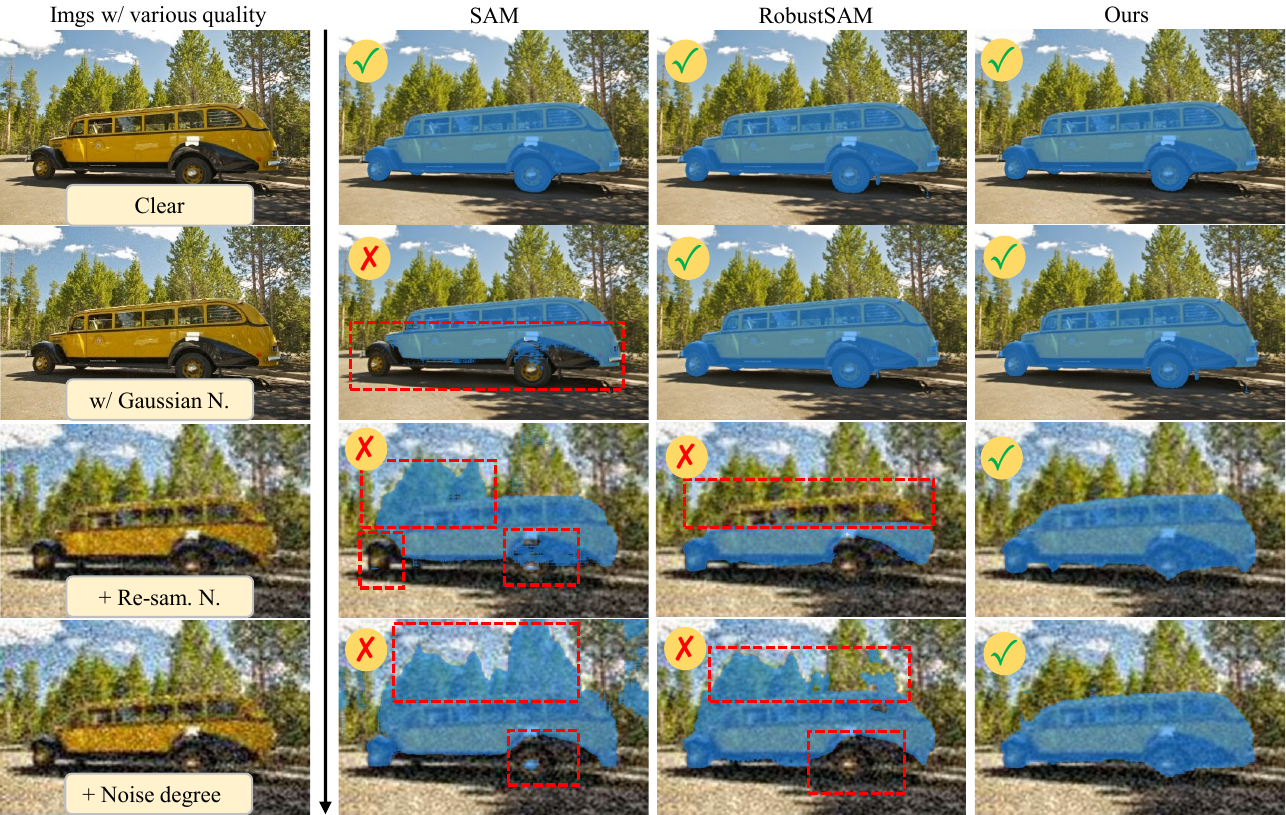}
   \caption{Qualitative results on low-quality images with varying degradation levels from an unseen dataset. To generate images with different degradation levels, we progressively added Gaussian Noise, Re-sampling Noise, and more severe Gaussian noise to an image. Results indicate that the baseline SAM \cite{sam} shows limited robustness to degradation. Although RobustSAM \cite{robustsam} retains some resilience against simpler degradations, it struggles with more complex and unfamiliar degradations. In contrast, our method consistently demonstrates strong robustness across images of varying quality.}
   \label{fig:comparison}
\end{figure}

\begin{figure*}[t]
  \centering
   \includegraphics[width=1\linewidth]{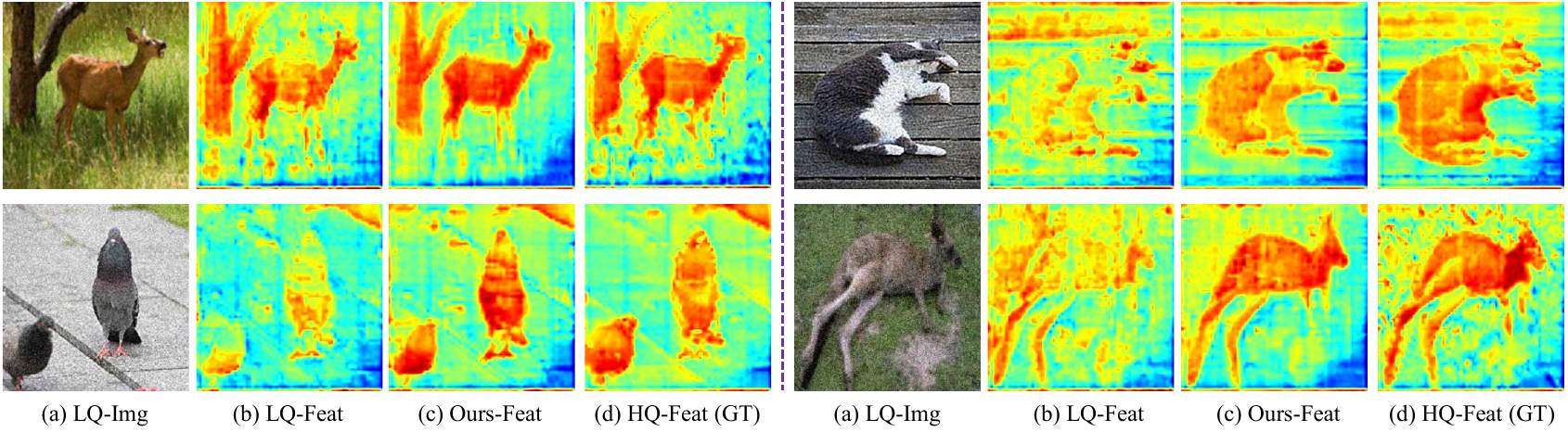}
   \caption{The visualization of latent features: (a) low-quality (LQ) images, (b) the SAM's latent features extracted from LQ images, which contain excessive noise and compromise the original representations, (c) enhanced representation by our GleSAM++, exhibiting more salient and well-preserved semantics and (d) the high-quality (HQ) features of the corresponding clear images, which are more salient than LQ ones.}
   \label{fig:lq-feat}
\end{figure*}

Despite their success, SAMs perform poorly on common low-quality images with various degradations, such as noise, blur, and compression artifacts \cite{sam-robustness1, sam-robustness2, sam-robustness3, robustness4}, which are often encountered in real-world scenarios \cite{image-noise, image-noise2, image-noise3}. 
To mitigate this gap, prior research \cite{robustsam, lwfa} has employed the discriminative robustness approach that trains models to learn degradation-invariant features. By leveraging consistent learning and aggressive data augmentation, these methods encourage similar outputs for clean and degraded image pairs.
However, they still face challenges in handling severely degraded low-quality images, where crucial information is irretrievably lost. 
As illustrated in Fig.~\ref{fig:comparison}, as degradations become more complex (\eg, combining various types of degradation or increasing the level of degradation), the existing methods \cite{sam, robustsam} struggle to accurately segment edges and complete target areas, leading to incorrect segmentation.
We analyze that it is caused by the limited feature representation for degraded images.
The visualizations in Fig.~\ref{fig:lq-feat} reveal that SAM's latent features from severely degraded images contain excessive noise, compromising the original representations and subsequently impacting the predictions of the decoder. Furthermore, the large gap between low-quality and high-quality features complicates consistency learning \cite{takd} in previous works \cite{robustsam}, as the discriminative model cannot recover details that no longer exist, hindering performance improvement. 
Thus, achieving high-quality latent feature representations and robust segmentation across varying image quality, especially for degraded images, remains an open and challenging problem.

These limitations call for a conceptual shift. The recently developed generative Diffusion Models (DM) \cite{ddpm, ddim}, especially the large-scale pre-trained Latent Diffusion Models (LDM) \cite{ldm} have demonstrated powerful content generation capabilities. Having been trained on internet-scale data \cite{laion}, LDM that proceeds diffusion and denoising in latent space, possesses a powerful representation prior, which can be well explored to enhance the latent representation of segmentation models.
This inspires us to \textbf{{take full advantage of the generative ability}} of pre-trained diffusion models and \textbf{{incorporate them into the latent space of SAMs}} to enhance low-quality features, thus promoting accurate segmentation in low-quality images.
Following this idea, rather than relying on discriminative approaches, we propose a new paradigm of \textbf{{Generative Latent space Enhancement}}, named \textbf{{GleSAM++}}, which reconstructs high-quality features (Fig. \ref{fig:lq-feat} (c)) in SAM's latent space through generative diffusion, thereby enabling accurate segmentation across images of varying quality.  
{Starting with low-quality features, high-quality representations are generated through single-step denoising.}
To integrate LDM generative knowledge, we incorporate a pre-trained U-Net from LDM with learnable LoRA layers \cite{lora} to align with segmentation-specific features.
Furthermore, to improve compatibility between the pre-trained diffusion model and the segmentation framework, we introduce Latent Space Alignment, including two effective techniques: \textbf{{Feature Distribution Alignment}} (FDA) and \textbf{{Channel Replication and Expansion}} (CRE). These techniques bridge feature distribution and structural alignment gaps between models.

While these components enable generative enhancement to operate reliably within SAM, the reconstruction process remains limited by the absence of explicit degradation-level guidance during the reconstruction process. 
This forces the denoising U-Net to implicitly learn the noise intensity while reconstructing features, increasing the learning difficulty and resulting in suboptimal feature enhancement.
To overcome this limitation, we further introduce a \textbf{{Degradation-aware Adaptive Enhancement}} (DAE) mechanism. 
It decouples the reconstruction process for arbitrary-quality features into two stages: degradation-level prediction and degradation-aware reconstruction. 
With this design, GleSAM++ explicitly predicts the degradation severity from the input features and dynamically modulates the denoising strength of the denoising process. Built upon SAMs, GleSAM++ leverages the generalization of pre-trained segmentation and diffusion models, with a few learnable parameters added, and can be efficiently trained within 16 hours on four GPUs.

In terms of data, systematic training and evaluation require samples covering a wide range of degradation levels. However, existing datasets lack sufficient diversity and comprehensiveness, posing a major challenge. To address this, we constructed \textbf{{LQSeg}} based on existing datasets \cite{thin, lvis, coco, msra, ecssd} to train and assess segmentation models on low-quality images.  
Compared with prior efforts \cite{robustsam}, LQSeg incorporates a broader diversity of degradation types than previous methods \cite{robustsam}, combining basic degradation models (\eg, noise and blur) to simulate complex and real-world noise \cite{realesrgan, bsrgan}. Furthermore, we introduce three degradation levels for a more comprehensive evaluation. We hope LQSeg will inspire the development of more robust segmentation models and contribute to future research.
Overall, our contributions are summarized as: 
\begin{itemize}
    \item We propose GleSAM++, a SAM-based framework that incorporates generative and adaptive latent space enhancement, to generalize across images of any quality. GleSAM++ demonstrates substantially improved robustness, particularly for low-quality images with diverse and complex degradations. 
    \item We develop Latent Space Alignment, including two effective techniques: Feature Distribution Alignment and Channel Replication and Expansion, to bridge feature distribution and structural gaps between the pre-trained LDM and SAM. 
    \item {We further introduce a degradation-aware adaptive enhancement mechanism, which enables the model to adaptively regulate the denoising strength in the latent space according to the predicted degradation level of the input features, thereby achieving substantially improved precision and robustness.}
    \item  We construct the LQSeg dataset, encompassing a diverse spectrum of degradation types and severities, to effectively train and evaluate the proposed model.
    \item Extensive experiments show that our method performs excellently on low-quality images with varying degrees of degradation while maintaining generalization to clear images. Additionally, our method achieves strong performance on unseen degradations, highlighting the adaptability of both our framework and dataset.
\end{itemize}

A preliminary version of this work (GleSAM) was published in CVPR2025 \cite{glesam}. This paper presents a substantial extension with the following key contributions.  
(1) First, we propose \textbf{GleSAM++, an enhanced version} with the key principle of decoupling the reconstruction process into degradation-level prediction and degradation-aware reconstruction. 
This design addresses a key limitation of the previous model, thereby \textbf{further improving performance across any-quality images}.
(2) Second, we \textbf{introduce a degradation-aware adaptive enhancement mechanism} that dynamically regulates the reconstruction intensity in the latent space based on degradation perception, \textbf{enabling a more targeted enhancement process}.
(3) Furthermore, we \textbf{conduct a more extensive experimental evaluation across a broader range of aspects and scenarios}, including robustness under diverse prompts and performance in real-world settings, thereby providing a more systematic validation of the model’s effectiveness.
(4) We also \textbf{provide additional analysis of the underlying factors contributing to GleSAM++’s effectiveness}, along with detailed algorithmic descriptions of both the training and inference processes.
(5) Finally, we \textbf{present comprehensive ablation studies and visual explanations}, offering deeper insights into the mechanisms behind GleSAM++.

\section{Related Work}
\label{sec:related}

\subsection{Segmentation on Low-Quality Images}
Executing robust segmentation across various scenarios is a critical issue.  Numerous studies \cite{benchmark-robustness, improve-robust, sam-robustness1, robustsam} have highlighted significant performance degradation in conventional segmentation models and foundational SAMs when confronted with low-quality images with degradation. 
Many related studies \cite{improve-robust, benchmark-robustness, lwfa, fifo, dense-gram} have been proposed to enhance the robustness of segmentation models against low-quality data.  For instance, URIE \cite{urie} enhances segmentation robustness under multiple image impairments by introducing classification-based constraints. QualNet \cite{QualNet} achieves quality-agnostic feature extraction through a reversible encoding scheme. Meanwhile, FIFO \cite{fifo} promotes the learning of fog-resilient features in segmentation frameworks via a fog-pass filtering mechanism.
These methods primarily consider a single type of degradation. 
{Segment Anything Model (SAM) \cite{sam} exhibits strong generalization ability due to large-scale pre-training on diverse data. However, this does not guarantee robustness under severe image degradations. Although diverse training data and augmentation improve invariance to moderate appearance variations, they cannot fully compensate for the loss of semantic cues caused by heavy noise, blur, or artifacts.}
Recently, RobustSAM \cite{robustsam} is introduced to enhance the robustness of the SAM against multiple image degradations through anti-degradation feature learning. However, its performance also struggles when dealing with complex degradations. The real-world image noise is often too complex to be modeled by a single degradation \cite{image-noise, image-noise2, image-noise3}. Therefore, robustly segmenting images of any quality remains challenging.

\subsection{Diffusion Models for Perception and Reconstruction Tasks}
Recently, diffusion models (DMs) \cite{ddpm, ldm, ddim, dmd, osdiff-vsd} have garnered significant attention in research, due to their powerful generation capabilities. Numerous studies \cite{segdiff, vpd, ddp, ldznet, diffusiondet, segrefiner, lessdm, diffbir, osediff} explore how to extend their applications to a broader range of tasks, such as detection, segmentation, and image reconstruction, \etc. 
For diffusion-based perception tasks, one category of methods \cite{ segdiff,diff-generalist, medsegdiff, medsegdiffv2} reformulate the perception tasks as progressive denoising from random noise. 
For example, DiffusionDet \cite{diffusiondet} and DiffusionInst \cite{diffusioninst} adapt the diffusion process to perform denoising in object boxes and mask filters.
Another route employs the pre-trained denoising UNet as a backbone for downstream perception tasks \cite{ddp, vpd, tadp, ldznet, ptdiffseg}. 
For example, VPD \cite{vpd}  passes the image through a pre-trained diffusion model and extracts intermediate features for task prediction. 
Diverging from these existing works, we preserve the original segmentation structure and fine-tune a generative diffusion to enhance the segmentation model’s latent representations for accurate segmentation of any quality images. 

For diffusion-based reconstruction tasks, DMs have shown powerful capabilities in image super-resolution (SR) \cite{stablesr, supir, diffbir}, deblurring \cite{deblurring-sr, msg-deblurring}, and low-light image enhancement \cite{lowlight-wdm, exposurediffusion} tasks, \etc.
It focuses on restoring degraded data, thus enabling the reconstruction of high-quality images with detailed semantics and realistic textures, even in scenarios characterized by severe and complex degradations.
For instance, StableSR \cite{stablesr} leverages
prior knowledge contained in pretrained text-to-image DMs for blind super-resolution. By utilizing a time-aware encoder, it achieves promising restoration results without modifying the pretrained synthesis model. DiffBIR \cite{diffbir} uses generative priors for SR, decoupling the restoration process into two stages.
SUPIR \cite{supir} further leverages multi-modal techniques and advanced generative priors.
However, these methods focus on improving visual quality for human perception, instead of improving the performance of downstream tasks. How to optimally apply these models to downstream tasks like segmentation is still unknown. 



\subsection{Segment Anything Model and Variants}
Segment Anything Models (SAMs) ~\cite{sam, sam2} have gained significant influence within the community due to their outstanding zero-shot segmentation capabilities. SAM \cite{sam} can interactively segment any object in an image using visual prompts such as points and bounding boxes. 
Most recently, the updated SAM2 \cite{sam2} has been released, showing improved segmentation accuracy and inference efficiency. 
Its robust generalization abilities have led to breakthroughs and new paradigms in various downstream tasks \cite{samhq, asam, samrs, samantic-aware-sam, sam-med, medsam2, 3dsam2, gaussian-grouping}. 
Although SAM is powerful, its performance decreases when facing complex scenarios, such as degraded images (including noise, blur, and adverse weather conditions) \cite{sam-robustness1, sam-robustness2, sam-robustness3, robustness4}, objects with intricate structures \cite{samhq} and camouflaged objects \cite{sam-struggle, sam-struggle2, sam-not-perfect}, which significantly hinders the real-world applications of SAM.
Enhancing SAM's capability in such challenging scenarios is a worthwhile research topic. 

Based on SAM, some improved variants have been proposed, which can be roughly categorized into two routes. 
One route  \cite{medical-sam-adapter, samadapter, sam-cod} involves using SAM for specific downstream tasks through domain-specific finetuning.
These efforts typically focus on improving SAM's performance on a specific task or dataset while sacrificing the model’s inherent generalization capabilities.
Another route \cite{samhq, robustsam, asam, mobilesam, vnssam} is to extend SAM's capabilities, preserving its strong generalization performance.
For example, HQ-SAM~\cite{samhq} has improved SAM's segmentation quality for objects with complex structures by adding adaptation layers while freezing SAM's original parameters. ASAM \cite{asam} enhances SAM's generalization capabilities through adversarial training.
Our approach follows the second path. However, diverging from these existing methods, our method focuses on improving the degradation robustness of SAM and introduces a generative latent space enhancement method within the SAM framework.

\begin{figure}[t]
  \centering
   \includegraphics[width=1.\linewidth]{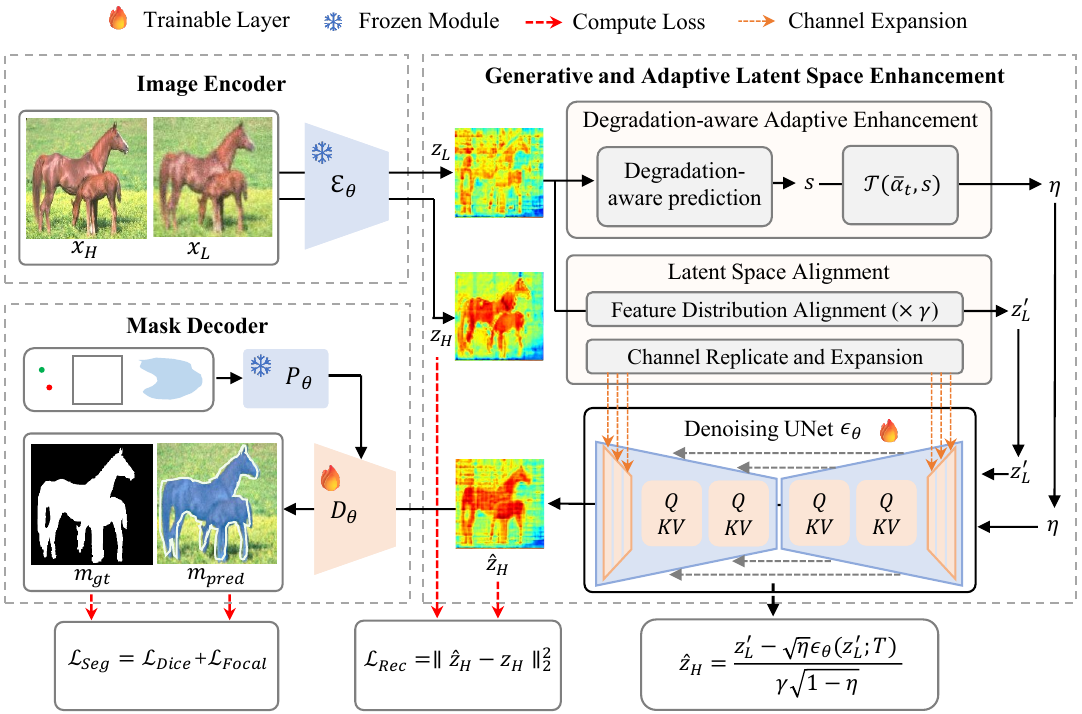}
   \caption{
   Given an input image, GleSAM++ performs accurate segmentation through image encoding, generative and adaptive latent space enhancement, and mask decoding. During training, with HQ-LQ image pairs as input, we adaptively reconstruct high-quality representations in the SAM's latent space by efficiently fine-tuning a generative denoising U-Net with LoRA layers. Degradation-aware adaptive enhancement is used to explicitly estimate the degradation level of the input features and uses this information to dynamically regulate the denoising strength. Latent space alignment is used to bridge the feature distribution and structural gaps between the pre-trained latent diffusion model and SAM.
   Subsequently, the decoder is fine-tuned with segmentation loss to align the enhanced latent representations. 
   Built upon SAMs, GleSAM++ inherits prompt-based segmentation and performs well on images of any quality.
   }
   \label{fig:pipeline}
\end{figure}

\section{GleSAM++: Generative Latent Space Enhancement for image Segmentation }  
\label{sec:method}
In the following, we explore how to improve SAM's robustness for low-quality images through generative latent space enhancement, thus enabling it to generalize across varying image qualities. The overall framework of the proposed \textbf{{GleSAM++}} is shown in Fig. \ref{fig:pipeline}.
To begin, in Sec. \ref{sec: fdm}, we first \textbf{model the method} of incorporating the generative priors of diffusion models (introduced in Sec. \ref{sec: preliminary}) to restore degraded latent features within SAM’s representation space. 
To ensure effective integration with the pre-trained diffusion backbone, we then develop \textbf{{Latent Space Alignment}} (Sec.~\ref{sec:lsa}), which addresses both distributional inconsistency and architectural mismatch between SAM features and diffusion latents via \textbf{{Feature Distribution Alignment}} and \textbf{{Channel Replicate and Expansion}}, which are detailed in Sec. \ref{sec: faw}, \ref{sec: REI}, respectively. 
While these components enable generative enhancement to function reliably within SAM, the reconstruction process remains constrained by its implicit handling of degradations with varying severity. The absence of explicit guidance complicates the learning process and often results in suboptimal reconstruction. To overcome this limitation, we further introduce a \textbf{{Degradation-aware Adaptive Enhancement}} mechanism in Sec. \ref{sec: glesam++}. 
By providing degradation-aware guidance, DAE allows the model to disentangle noise estimation from degradation severity, thereby reducing learning complexity and enabling more accurate reconstructions
Finally, the overall training method and in-depth discussions are outlined in Sec. \ref{sec: ft} and Sec. \ref{sec:model_analysis}. 


\subsection{Preliminary: Latent Diffusion Model}
\label{sec: preliminary}
Diffusion models (DMs) \cite{ddpm, ddim, ldm} are generative models parametrized by a Markov chain and composed of forward and backward processes. They progressively add noise to the original pixel space in the forward process, and then learn to reverse this process by predicting and removing the noise. Formally, in DMs, the forward noise process iteratively adds Gaussian noise with variance $\beta_t \in (0,\mathrm{I})$ to the variable $z$. The sample at each time point is defined as: 
\begin{equation}
    z_t =\sqrt{\overline{\alpha}_{t}}z + \sqrt{1-\overline{\alpha}_t}\epsilon,
    \label{eq:1}
\end{equation}
where $\alpha_t = 1-\beta_t$, $\overline{\alpha}_{t}=\prod_{s=1}^{t}\alpha_s$, and $\epsilon \in \mathcal{N}(0, \mathrm{I})$. 
While the inverse diffusion process is modeled by applying a neural network $\epsilon_{\theta}(z_t, t)$ to predict the noise $\hat{\epsilon}$ and recover the original input $z$. 
Latent Diffusion Models (LDMs) perform this generative process in a compressed latent space using a pre-trained Variational Autoencoder (VAE) \cite{vae}. The encoder maps high-dimensional images into a lower-dimensional latent representation, while the decoder reconstructs the enhanced signal at the original spatial resolution, enabling more efficient computations in the training and inference phases.

\subsection{Modeling: Latent Denoising Diffusion in Segmentation}
\label{sec: fdm}    

Inspired by the principles of LDMs, we extend the generative denoising process to latent space enhancement of SAM. 
Specifically, we aim to leverage the latent diffusion mechanism to reconstruct high-quality segmentation representations from their degraded counterparts.
Let's denote $\mathcal{E}_{\theta}$ and $\mathcal{D}_{\theta}$ the segmentation encoder and decoder of SAMs, respectively. 
As shown in Fig. \ref{fig:pipeline}, given a pair of high-quality (HQ) and low-quality (LQ) images $\{x_H, x_L\}$  as inputs, the encoder produces corresponding latent features $\{z_H, z_L\}$:
\begin{equation}
    z_H, z_L = \mathcal{E}_{\theta}(\{x_H, x_L\}).
\end{equation}
Here, $z_L$ can be considered to be a noisy version of $z_H$, which still retains information to reconstruct a high-quality feature. 
Instead of the complex multi-step denoising from random noise, we start directly from $z_L$ and forward with a single denoising step. Specifically, based on Eq. \ref{eq:1}, the denoised latent variable $z$ can be directly predicted from the model's predicted noise $\hat{\epsilon}$, as: 
\begin{equation}
    \hat{z} = \frac{z_t-\sqrt{1-\overline{\alpha}_t}\hat{\epsilon}}{\sqrt{\overline{\alpha}_t}},
\end{equation}
where $\hat{\epsilon}$ is the prediction of the network $\epsilon_{\theta}$ with given $z_t$ and t: $\hat{\epsilon}=\epsilon_{\theta}(z_t;t)$.
We re-parameterize the above generative denoising process to adapt low-quality latent space enhancement in segmentation, as:
\begin{equation}
    \hat{z}_H = \mathrm{GLE}(z_L)= \frac{z_L-\sqrt{1-\overline{\alpha}_T}\epsilon_{\theta}(z_L; T)}{\sqrt{\overline{\alpha}_T}},
    \label{eq: denoise}
\end{equation}
where we consider the low-quality feature $x_L$ as the noised feature and perform one-step denoising to reconstruct it efficiently.
The denoised output $\hat{z}_H$ is expected to be more closely to the features extracted from clear images $z_H$. 
This single-step process significantly reduces computational overhead, making it more efficient when applied to segmentation models. 
Finally, the refined latent $\hat{z}_H$ is passed through the mask decoder to predict segmentation masks, as: 
\begin{equation}
    m_p=\mathcal{D}_{\theta}(\hat{z}_H, \mathcal{P}_{\theta}(p)),
\end{equation}
where $\mathcal{P}_{\theta}(\cdot)$ and $p$ are the prompt encoder and visual prompts, respectively. 

\subsection{Latent Space Alignment}
\label{sec:lsa}
In our framework, the pre-trained U-Net from the LDM \cite{ldm} serves as the generative denoising backbone.
The proposed method in Sec. \ref{sec: fdm} establishes a foundation for enhancing SAM with generative latent denoising. 
While it provides a promising direction, practical deployment reveals critical incompatibilities between segmentation features and the pre-trained U-Net backbone, leading to several technical issues
These include (1) a distribution mismatch between VAE-derived latent variables and segmentation embeddings, and (2) a channel dimensionality gap between the U-Net design and SAM’s feature space.
To overcome these limitations, we introduce two techniques: Feature Distribution Alignment and Channel Replication–Expansion.

\begin{wrapfigure}{r}{0.28\textwidth} 
    \vspace{-10pt}
    \centering
        \includegraphics[width=\linewidth]{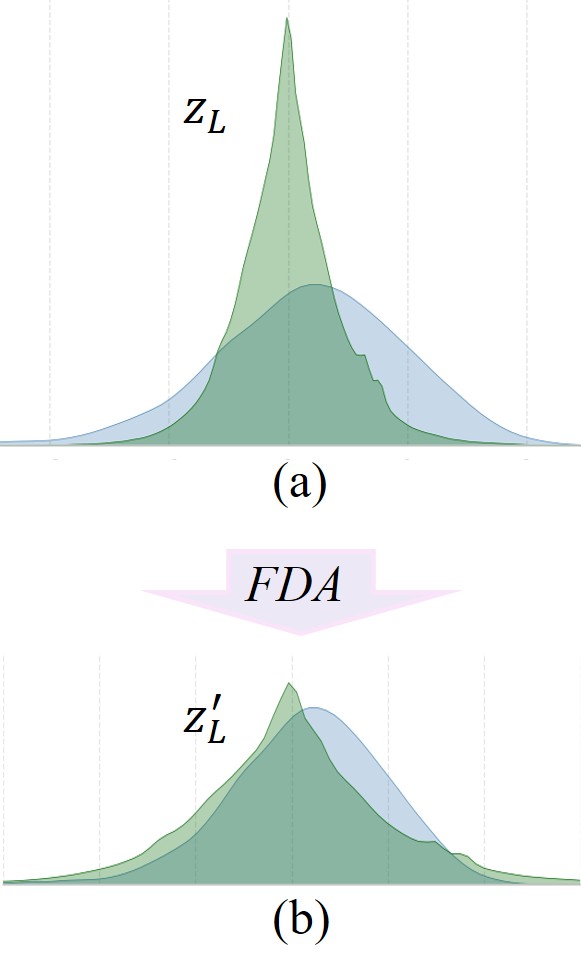}
        \caption{Aligning feature distributions before (a) and after (b) FDA.}
        \label{fig:fda}
    \vspace{-30pt}
\end{wrapfigure}

\subsubsection{Feature Distribution Alignment}
\label{sec: faw}
Firstly, there is a distribution gap (as shown in Fig. \ref{fig:fda} (a)) between the two spaces and directly feeding segmentation features into the U-Net may prevent it from fully exerting its denoising capabilities.
To address this gap, we introduce a Feature Distribution Alignment (FDA) technique. Specifically, we add an adaptation weight $\gamma$ to scale the segmentation features, adjusting their variance to align more closely with the VAE's latent space (Fig. \ref{fig:fda} (b)).
This adjustment ensures that the features are compatible with U-Net’s optimal input space, improving the robustness and accuracy of the semantic interpretation and enhancing the denoising capability.
The LQ feature denoising process in Eq. \ref{eq: denoise} can be updated as:
\begin{equation}
    \hat{z}_H = \mathrm{GLE}(z_L)= \frac{\gamma z_L-\sqrt{1-\overline{\alpha}_T}\epsilon_{\theta}(\gamma z_L; T)}{\gamma \sqrt{\overline{\alpha}_T}},
    \label{eq: denoise-2}
\end{equation}
where we divide by $\gamma$ to restore its original distribution.
We experimentally verified in Sec. \ref{sec: ablation-modules} that this simple yet principled alignment significantly improves the U-Net’s ability to reconstruct high-quality features from degraded segmentation inputs.

\subsubsection{Channel Expansion for Head-tail Layers}
\label{sec: REI}
Another technical issue arises from the channel mismatch of the head and tail layers between the pre-trained U-Net and the segmentation features.  The LDM’s U-Net is designed to process inputs and outputs of dimension $h \times w \times 4$, reflecting the 4-channel latent representation produced by its VAE encoder. In contrast, SAM’s latent space has substantially higher dimensionality, namely $h \times w \times 256$. Directly feeding these representations into the U-Net is therefore infeasible. 

{
A straightforward solution is to learn new projection layers or newly initialized head/tail layers to bridge this mismatch. However, such strategies are suboptimal not only empirically (as shown in Sec. \ref{sec: ablation-cre}) but also from the perspective of compatibility with the pretrained diffusion model. 
The head and tail layers of the U-Net are jointly optimized with the internal denoising backbone during large-scale pretraining, and therefore form a coupled parameterization with the whole generative model. 
If these layers are replaced by newly initialized mappings, the transformed SAM features may exhibit activation statistics that deviate substantially from those expected by the pretrained U-Net, making the pretrained generative prior harder to transfer effectively and weakening the generalization ability of the backbone.}
To overcome this, we propose a Channel Replication and Expansion (CRE) strategy. Concretely, we replicate and concatenate the pre-trained head and tail weights of the U-Net, thereby expanding the input and output channels from 4 to 256 without altering the original parameter distribution.  During training, the replicated head and tail layers are kept frozen to preserve their pre-trained generalization, while learnable LoRA \cite{lora} layers are added to adapt to segmentation features. This design not only ensures compatibility with SAM’s latent space but also minimizes the number of learnable parameters, enabling efficient fine-tuning without sacrificing the robustness of the pre-trained generative backbone.

\subsection{Degradation-aware Adaptive Enhancement}
\label{sec: glesam++}
As outlined in the above sections, our framework successfully integrates generative enhancement into SAM by leveraging the strong generative priors of pre-trained diffusion models to reconstruct degraded latent features. However, it still remains limited by its need to implicitly accommodate degradations of different severity levels.  As a result, the model is forced to implicitly fit a complex noise distribution that spans conditions from mild noise to severe artifacts. 
This lack of explicit guidance complicates the learning process and leads to suboptimal results.
To address this limitation, we further consider \textbf{decoupling the reconstruction process for arbitrary-quality features into two stages: a degradation-aware prediction stage and a degradation-aware reconstruction stage}. 
By introducing degradation-aware guidance, the denoising UNet can focus on learning the noise distribution itself, rather than implicitly predicting different noise intensities at the same time. This thereby reduces the complexity of the learning process. Note that predicting noise intensity (\eg, regressing a scalar) is much simpler than predicting high-dimensional noise.
Following this idea, we introduce the Degradation-aware Adaptive Enhancement (DAE) mechanism. 
It explicitly estimates the degradation level of the input features and uses this information to dynamically regulate the denoising strength in the latent space.  


\begin{wrapfigure}{r}{0.43\textwidth} 
    \vspace{-20pt}
    \centering
        \includegraphics[width=\linewidth]{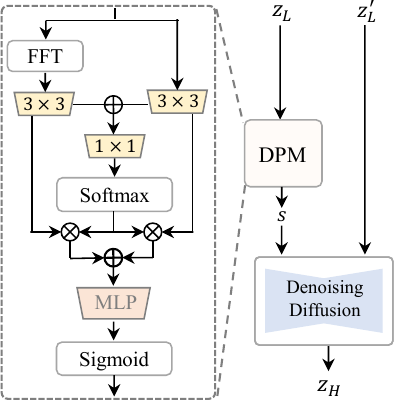}
        \caption{Details of the degradation-aware prediction module.}
        \label{fig:DAE_module}
    \vspace{-20pt}
\end{wrapfigure}

\subsubsection{Degradation-aware Prediction Module} 
The DAE mechanism must be driven by an accurate signal that quantifies the input's degradation. 
{While traditional Image Quality Assessment (IQA) methods \cite{maniqa-iqa, arniqa-iqa, depicting-iqa, vlc-iqa} exist, they are fundamentally incompatible with our framework. Primarily, IQA metrics are optimized to predict human perceptual quality, which correlates poorly with the feature-level distortions that directly impact a deep network's performance.}
To this end, we introduce a simple, learnable degradation-aware prediction module $\text{DPM}(\cdot)$ to learn a task-specific degradation score $s \in [0,1]$ tailored for segmentation task. 
The design of DPM is motivated by the observation that degradation artifacts manifest in two complementary domains: the spatial domain (\eg, compression blockiness) and the frequency domain (\eg, blur resulting in loss of high frequencies).

\begin{wrapfigure}{r}{0.5\textwidth} 
    \centering
        \includegraphics[width=\linewidth]{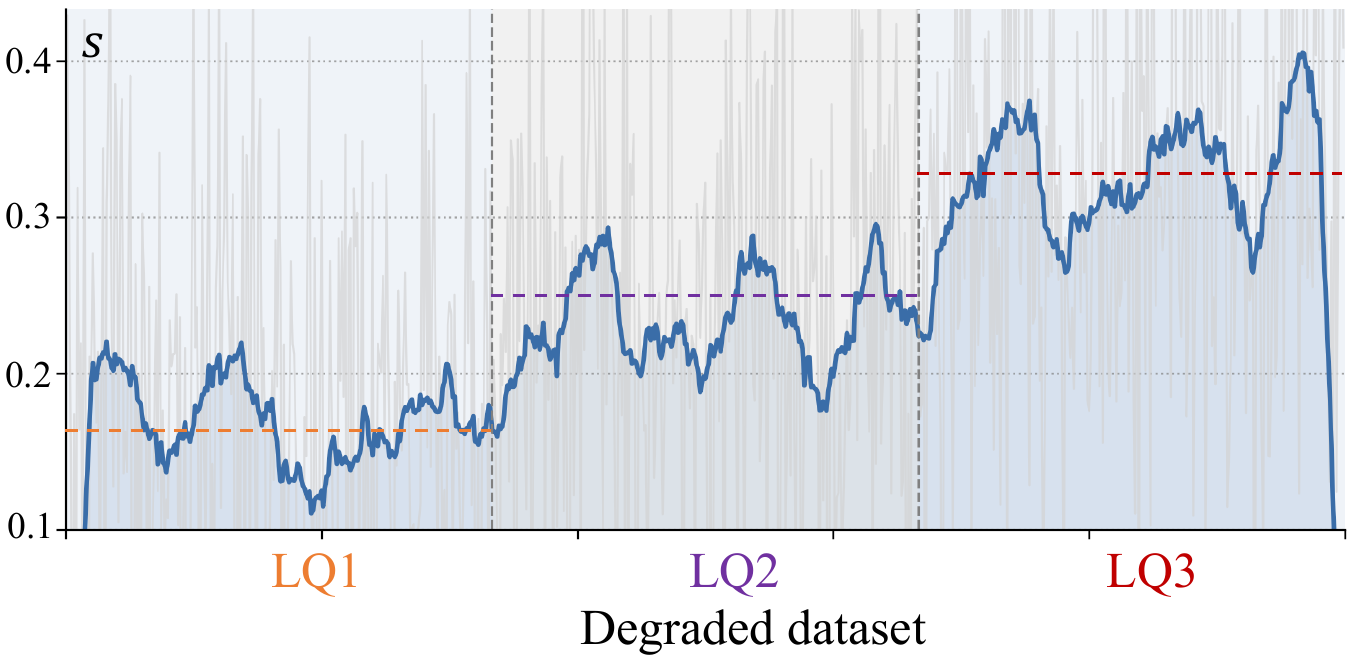}
        \caption{Predicted degradation scores across datasets with different degradation levels.}
        \label{fig:score}
    \vspace{-20pt}
\end{wrapfigure}

As illustrated in Fig. \ref{fig:DAE_module}, the input feature $z_L$ is first processed through a $3 \times 3$ depth-wise convolution to capture localized structural distortions. In parallel, a frequency branch transforms $z_L$ into its amplitude spectrum using a Fast Fourier Transform (FFT) layer, followed by a convolution layer that extracts frequency corruption patterns. The two branches are then fused via element-wise addition to form a hybrid descriptor. Then, a $1 \times 1$ convolution is used to compress the hybrid feature into a single-channel map, which is normalized by a spatial softmax operation. The resulting attention weights are applied to both spatial and frequency features, producing compact descriptors that are aggregated and passed through an MLP layer with a sigmoid activation. The final output is a continuous degradation score $s$, where larger values correspond to more severe degradation. 
In Fig. \ref{fig:score}, we depict the predicted degradation scores $s$ across datasets with varying degradation levels (LQ1–LQ3). 
It can be seen that the predicted $s$ increase progressively from LQ1 to LQ3, aligning well with the predefined degradation levels, which validates that our simple DPM can effectively perceive and quantify degradation information.

\subsubsection{Degradation-aware Latent Space Reconstruction} 
The predicted $s$ functions as an explicit control signal for the generative enhancement process. 
In diffusion-based denoising, the parameter $\overline{\alpha}_t$ controls the signal-to-noise ratio at timestep $t$, as shown in Eq. \ref{eq: denoise-2}. Our framework adaptively modulates this parameter based on $s$, effectively mapping inputs of varying quality to different positions on the diffusion trajectory.

Rather than directly replacing $\overline{\alpha}_t$ with $s$, we perform a linear interpolation between a pre-defined minimum and maximum noise level: 
\begin{equation}
    \eta = \mathcal{T}(1-\overline{\alpha}_t, s) = \overline{\beta}_{t_{min}} + s \cdot(\overline{\beta}_{t_{max}} - \overline{\beta}_{t_{min}}),
\end{equation}
where we set $\overline{\beta}_t = 1 - \overline{\alpha}_t$ for simplicity.
$\overline{\beta}_{t_{\min}}$ and $\overline{\beta}_{t_{\max}}$ correspond minimum and maximum noise level respectively. 
This design constrains the adaptive noise ratio to a valid and stable range, thereby preventing pathological values that could destabilize the denoising process. It significantly enhances the stability of the end-to-end training process.  The resulting parameter $\eta$ is then integrated into the generative latent enhancement step. Thus Eq. \ref{eq: denoise-2} is reformulated as:
\begin{equation}
    \hat{z}_H = \mathrm{GLE}(z_L, \eta)= \frac{\gamma z_L-\sqrt{\eta}\epsilon_{\theta}(\gamma z_L; T)}{\gamma \sqrt{1-\eta}}.
    \label{eq: denoise-}
\end{equation}

\begin{center}
\begin{algorithm}
\caption{\textbf{Training Scheme of GleSAM++}}
\label{al:train}
\begin{algorithmic}[1]
\State \textbf{Input:} Training dataset $\mathcal{S}$, pretrained SAM including image encoder $\mathcal{E}_{\theta}$, prompt encoder $\mathcal{P}_{\theta}$, and mask decoder $\mathcal{D}_{\theta}$. Pretrained U-Net $\epsilon_{\theta}$ with learnable LoRA layers, fine-tuning iteration $N_1, N_2$. \\
\textit{/*               Fine-tuning U-Net                */}
\For{$i \leftarrow 1$ to $N_1$}
    \State Sample $x_H, x_L$ from $\mathcal{S}$
    \State \textit{/* Network forward */}
    \State $[z_H, z_L] \leftarrow \mathcal{E}_{\theta}([x_H, x_L])$
    \State $s \leftarrow \mathrm{DPM}(z_L)$
    \State $\hat{z}_H \leftarrow \mathrm{GLE}(z_L, s)$
    \State \textit{/* Compute reconstructive loss */}
    \State $\mathcal{L}_{\mathrm{Rec}} = \mathcal{L}_{\text{MSE}}(\hat{z}_H, z_H)$
    \State \textit{/* Network parameter update */}
    \State Update learnable parameters with $\mathcal{L}_{\mathrm{Rec}}$
\EndFor \\
\textit{/* Fine-tuning Decoder */}
\For{$i \leftarrow 1$ to $N_2$}
    \State Sample $\hat{z}_{H}, m_g$ from $\mathcal{S}$
    \State \textit{/* Network forward */}
    \State Sample prompts $p$ from $m_g$ 
    \State $m_p \leftarrow \mathcal{D}_{\theta}(\hat{z}_{H}, \mathcal{P}_{\theta}(p))$
    \State \textit{/* Compute segmentation loss */}
    \State $\mathcal{L}_{\mathrm{Seg}} = \mathcal{L}_{\text{Dice}}(m_p, m_g) + \mathcal{L}_{\mathrm{Focal}}(m_p, m_g)$
    \State \textit{/* Network parameter update */}
    \State Update learnable parameters with $\mathcal{L}_{\mathrm{Seg}}$
\EndFor
\State \textbf{Output:} Fine-tuned U-Net $\epsilon_{\theta}$ and mask decoder $\mathcal{D}_{\theta}$
\end{algorithmic}
\end{algorithm}
\end{center}

For high-quality images with a low score $s$, $\eta$ will be small, resulting in a gentle denoising effect that preserves details. For severely degraded images with a high score $s$, $\eta$ will be large, leading to a much stronger denoising effect to reconstruct the underlying features. 
By conditioning the denoising process on the predicted degradation severity, the proposed DAE module provides an adaptive and principled mechanism for targeted feature restoration. This design significantly enhances GleSAM++’s robustness and generalization across a broad spectrum of degradation conditions.



\begin{algorithm}
\caption{\textbf{Inference Scheme of GleSAM++}}
\label{al:inference}
\begin{algorithmic}[1]
\State \textbf{Input:} Low-quality image $x_L$ and corresponding prompt $p$, Pretrained image encoder $\mathcal{E}_{\theta}$ and prompt encoder $\mathcal{P}_{\theta}$. Fine-tuned mask decoder $\mathcal{D}_{\theta}$ and U-Net $\epsilon_{\theta}$.
\State \textit{/* Image encoding */}
\State $z_L \leftarrow \mathcal{E}_{\theta}(x_L)$
\State \textit{/* Generative latent space enhancement */}
\State $s \leftarrow \mathrm{DPM}(z_L)$
\State $\hat{z}_H \leftarrow \mathrm{GLE}(z_L, s)$
\State \textit{/* Mask decoding */}
\State $m_p \leftarrow D_{\theta}(\hat{z}_{H}, \mathcal{P}_{\theta}(p))$
\State \textbf{Output:} Predicted mask $m_p$.
\end{algorithmic}
\end{algorithm}

\subsection{Training Method}
\label{sec: ft}
{We employ a two-stage fine-tuning strategy for stable and efficient optimization. In the first stage, we adapt the denoising U-Net to restore degraded semantic features in a parameter-efficient manner, while keeping the SAM image encoder frozen. In the second stage, we fine-tune the mask decoder using the restored features for accurate segmentation. This staged design does not increase practical training complexity, since only a limited number of parameters are optimized. Instead, it improves optimization stability by decoupling feature restoration from segmentation learning. }
Detailed training and inference schemes are shown in Algorithm \ref{al:train} and \ref{al:inference}.

\noindent\textbf{U-Net finetuning.} To adapt the pre-trained U-Net to the segmentation framework while preserving its inherent generalization ability, we employ the LoRA \cite{lora} scheme to fine-tune all the attention layers in the U-Net. During this step, we freeze the pre-trained image encoder and U-Net layers and only fine-tune the added LoRA layers. The estimated feature is compared with the corresponding HQ feature $z_H$ by a reconstruction loss, as:
\begin{equation}
    \mathcal{L}_{\mathrm{Rec}} = \mathcal{L}_{\mathrm{MSE}}(\mathrm{GLE}(z_L, \eta), z_H).
\end{equation}
This step significantly enhances performance without fine-tuning SAM’s parameters.

\noindent\textbf{Decoder finetuning.} Next, we use the reconstructed high-quality features to fine-tune the mask decoder for more precise segmentation. Our experiments demonstrate that fine-tuning either the entire decoder or only the output tokens with these features further improves segmentation accuracy while maintaining generalization on clear images.
Focal Loss and Dice Loss are employed as segmentation loss, as:
\begin{equation}
    \mathcal{L}_{\mathrm{Seg}} = \mathcal{L}_{\mathrm{Dice}}(m_p, m_g) + \mathcal{L}_{\mathrm{Focal}}(m_p, m_g),
\end{equation}
where $m_p$ and $m_g$ indicate predicted and ground-truth masks, respectively.

\subsection{Analysis of GleSAM++}
\label{sec:model_analysis}
In this section, we discuss the key principles that enable the effectiveness of GleSAM++. Its robust performance arises from three core concepts central to our approach.
(1) \textbf{Powerful generative priors}: GleSAM++ leverages the powerful generative priors embedded in the pre-trained LDMs. Trained on massive internet-scale image datasets, LDMs encode a profound understanding of natural image statistics.  
Compared with previous discriminative approaches \cite{robustsam, lwfa}, the powerful generative priors in our generative enhancement method can effectively reconstruct clean and salient features, even when the input representations are severely corrupted by degradation.
(2) \textbf{Modeling a generalizable transformation}: We frame feature enhancement as a diffusion denoising process. 
Our core insight is to view the features from a degraded image as a noisy version of the clean features. 
This noise is a general concept representing any feature corruption, such as blur, compression, or their mixture. 
Consequently, the reverse diffusion process learned by the LDM models a highly generalizable transformation. It is capable of mapping samples from these arbitrary corrupted distributions back to the clean feature manifold. 
By formulating our enhancement this way, GleSAM++ does not overfit to specific degradation types. Instead, it learns a universal restoration mechanism. This explains its robust performance across a diverse spectrum of degradations, including those unseen during training, as has been validated comprehensively by our experiments.
(3) \textbf{Task Decoupling for Conditional Enhancement}: To alleviate the learning burden, GleSAM++ further divides the feature reconstruction process into two specialized sub-tasks: a degradation prediction stage and a conditional reconstruction stage. This division of labor significantly reduces the learning burden on the reconstruction module. It no longer needs to implicitly model the degradation level and can instead dedicate its full capacity to the restoration process itself, leading to a more precise and higher-fidelity feature enhancement.


\section{Low-Quality Image Segmentation Dataset}
\label{sec:lq-dataset}
To effectively train and evaluate the robust model, we construct a comprehensive low-quality image segmentation dataset, dubbed \textbf{LQSeg}, that encompasses more complex and multi-level degradations, rather than relying on a single type of degradation. The dataset is composed of images from several existing datasets with our synthesized degradations.
In this section, we first introduce a multi-level degradation process of low-quality images and then detail the dataset composition.

\begin{figure*}[t]
  \centering
   \includegraphics[width=1\linewidth]{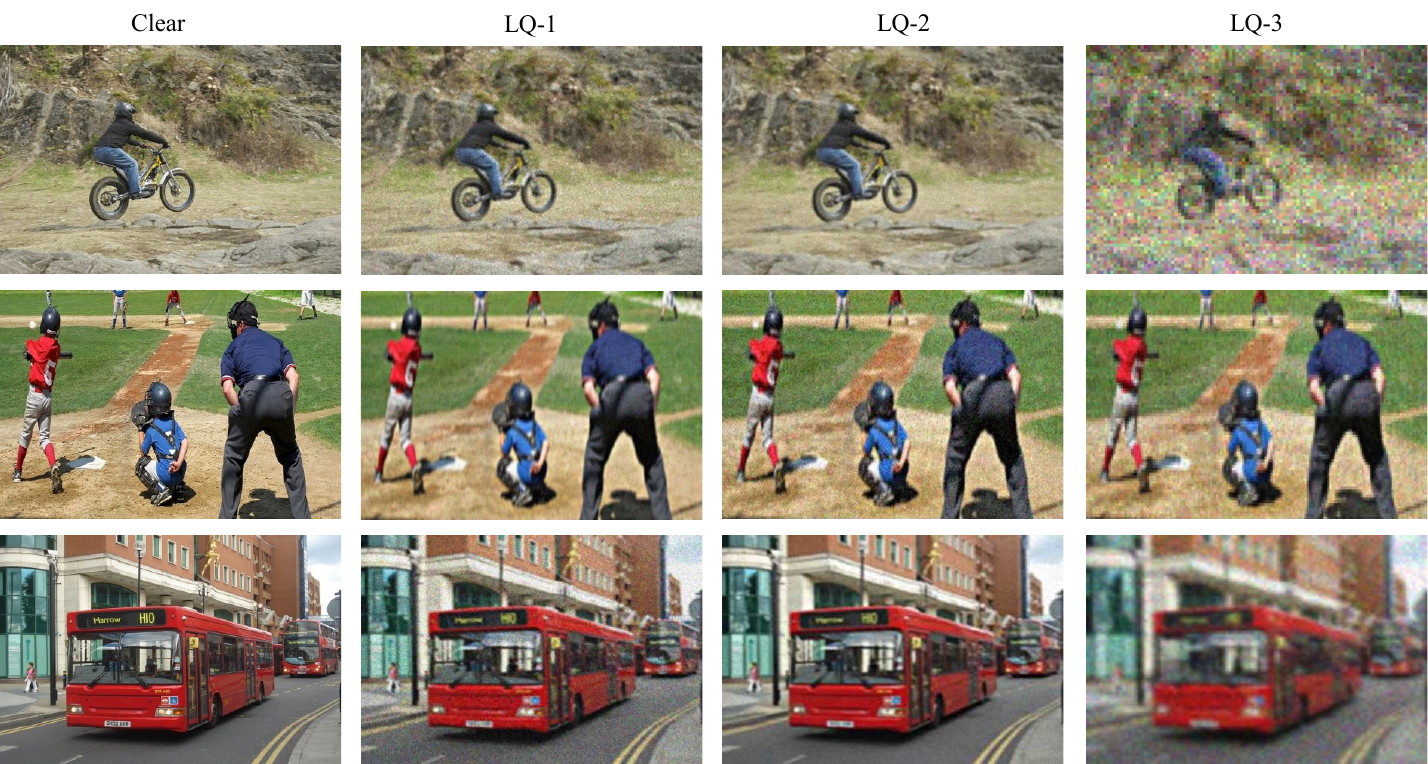}
   \caption{Examples from the LQ-Seg dataset illustrating images with varying levels of synthetic degradation: LQ-1, LQ-2, and LQ-3. These samples showcase the progressive quality deterioration used for evaluating the robustness of segmentation models.}
   \label{fig:sm-lqimg}
\end{figure*}

\subsection{{Multi-level Degradation Modeling}}
To model a more practical and complex degradation process, inspired by the previous work in image reconstruction \cite{realesrgan, bsrgan}, we utilize a mixed degradation method. 
Specifically, the degraded process is modeled as the random combination of the four common degradation models, including \textit{Blur}, \textit{Random Resize}, \textit{Noise}, and \textit{JEPG Compression}. 
Each degradation model encompasses various types, such as Gaussian and Poisson noise for \textit{Noise}, ensuring the diversity of the degradation process. 

Specifically, 1) For {\textbf{Blur}} degradation, it is typically modeled as a convolution with a blur kernel. We randomly choose Gaussian kernels, generalized Gaussian kernels, and plateau-shaped kernels, with preset probability, kernel size, and standard deviation. 2) For {\textbf{Random Resize}} operation, we consider both upsampling and downsampling operations with preset resize scales and randomly selected resize algorithms (\textit{i.e.}, bilinear interpolation, bicubic interpolation, and area resize). The randomness benefits include more diverse and complex resize effects. 3) For {\textbf{Noise}} degradation, we consider two commonly used noise types: Gaussian noises and Poisson noises. Gaussian noise has a probability density function equal to that of the Gaussian distribution. The noise intensity is controlled by the standard deviation of the Gaussian distribution. Poisson noise follows the Poisson distribution, which is usually used to approximately model the sensor noise caused by statistical quantum fluctuations, that is, variation in the number of photos sensed at a given exposure level \cite{realesrgan}. 4) For {\textbf{JPEG Compression}} operation, we use the off-the-shelf algorithms \cite{jpeg}, with a preset quality factor range.  

To enrich the granularity of degradation,  we employ multi-level degradation by adjusting the downsampling rates. We employed three different resize rates, \textit{i.e.}, [1, 2, 4], which correspond to three degradation levels from slight to severe: LQ-1, LQ-2, and LQ-3. Fig. \ref{fig:sm-lqimg} shows sample images with varying levels of synthetic degradation from the LQ-Seg dataset, demonstrating the diversity of degradation.


\begin{table}[t] \small
\caption{The source data composition of the LQSeg dataset.}
\centering
\renewcommand\arraystretch{1}
\setlength{\tabcolsep}{3.5 mm}{
\begin{tabular}{lllc}
\toprule[0.8pt]
\multicolumn{4}{c}{\textbf{\textit{Low-Quality Image Segmentation Dataset}}} \\
Set Type & Evaluation Category & Source data & Num. \\
\toprule[0.5 pt]
\multirow{4}{*}{{Training Set}} & \multirow{4}{*}{--} & LVIS \cite{lvis} & 15,000 \\
 & & ThinObject-5K \cite{thin} & 4748 \\
 & & MSRA10K \cite{msra} & 10,000 \\
 \cmidrule(l){3-4}
 & & \textbf{Total} & \textbf{29748} \\
\midrule[0.5pt]
\multirow{6}{*}{{Evaluation Set}} & \multirow{2}{*}{{In-distribution}} & ThinObject-5K (test set) \cite{thin} & 500 \\
 & & LVIS (test set) \cite{lvis} & 2,000 \\   \cmidrule(l){3-4}
 & \multirow{2}{*}{{Cross-dataset}} & ECSSD \cite{ecssd} & 1,000 \\ 
 & & COCO-val \cite{coco} & {1,392} \\ \cmidrule(l){3-4}
 & \multirow{2}{*}{{Out-of-distribution}} & Robust-Seg \cite{robustsam} & 1,000  \\
 &  & BDD-10k \cite{bdd100k} & 1,000 \\
\toprule[0.8pt]
\end{tabular}}
\label{tab:datasets}
\vspace{-5pt}
\end{table}

\subsection{Dataset Composition}
Based on the above multi-level degradation model, we construct \textbf{{LQSeg}} to train our model and evaluate the segmentation performance on different levels of low-quality images.
The images in LQSeg are sourced from several well-known existing datasets in the community with our synthesized degradation. The dataset details are shown in Tab. \ref{tab:datasets}.
In detail, for the \textbf{{training set}}, we utilize the entire training sets of LVIS \cite{lvis}, ThinObject-5K \cite{thin}, and MSRA10K \cite{msra} as the source data and procedurally  
synthesize corresponding low-quality images. 
{The \textbf{{evaluation set}} is organized into three progressively challenging protocols to comprehensively assess both segmentation accuracy and generalization ability:
\textbf{1) In-distribution evaluation.} We use the test splits of ThinObject-5K \cite{thin} and LVIS \cite{lvis}, whose source datasets are involved in training.
\textbf{2) Cross-dataset evaluation.} We include ECSSD \cite{ecssd} and a de-overlapped subset of COCO-val \cite{coco}. Since LVIS \cite{lvis} annotations are built upon COCO images, we explicitly remove all COCO-val images that appear in the LVIS training set to eliminate any image-level data leakage. The remaining COCO-val images share no overlap with our training data.
\textbf{3) Out-of-distribution evaluation.} We further use RobustSeg \cite{robustsam} and BDD-10K \cite{bdd100k} to evaluate the model under degradation types not included in the training set.
BDD-10K includes a variety of real-world degradations such as low light, blur, rain, and snow, which ensures a thorough evaluation of GleSAM++'s robustness in realistic scenarios.
For each source image in protocols 1) and 2), we systematically generate three levels of degraded images to thoroughly assess the model's robustness.}

\section{Experiment}
We conduct extensive experiments to verify our method across images of varying quality. All proposed techniques can be applied to SAM and SAM2, referred to as GleSAM++ and GleSAM2++. In practice, our models perform well on low-quality (Tab. \ref{tab:thin}, \ref{tab:unseen-sets}) and they generalize effectively to unseen degradations (Tab. \ref{tab:bdd}). 

\subsection{Experimental Setup}
\subsubsection{Implement details}
Built upon SAMs, GleSAM++/GleSAM2++ inherits prompt-based segmentation. During training, we utilize random points or the bounding box as prompts, which are encoded into prompt vectors by the frozen prompt encoder and then fed into the decoder. Our model is trained using the AdamW \cite{adamw} optimizer with the learning rate of $2\times10^{-4}$ and batch size of 4. 
The pre-trained U-Net in Stable Diffusion (SD) 2.1-base \cite{ldm} is adopted as the denoising backbone.
Our approach can be efficiently trained on 4$\times$ RTX 4090 GPUs within approximately 16 hours, during which we fine-tune the U-Net for 50K iterations and the decoder for only 20K iterations. 
During the inference, our methods follow the interactive pipeline as SAM, ensuring compatibility. 
For point-prompted evaluation, we randomly sample several points from the ground truth masks and use them as the input prompt. In our experiments, the number of random points is set to 3 by default. 
Additionally, we also use ground-truth boxes and noise boxes as prompts. 
For box-prompted evaluation, we use the ground truth mask to generate the bounding box and input it as the box prompt. For noise-box-prompted evaluation, the noise-box is generated by adding noise to the GT box as the prompt input, following \cite{dndetr}.
In our experiments, the noise scale is set to 0.2 by default.

\subsubsection{Comparison baselines} 
We compare our method with a set of comparison baselines to quantify the performance gains.
1) SAM \cite{sam} and SAM2 \cite{sam2}, serving as the standard foundation segmentation baselines. {While SAM2 is originally designed for video segmentation, our use of SAM2 here is intended to verify the compatibility of the proposed latent enhancement framework with a stronger backbone at the frame representation level. The current evaluation therefore focuses on degraded-image segmentation, and does not constitute a systematic study of temporal consistency under degraded video inputs.}
2) Besides SAMs, we also compare with the RobustSAM \cite{robustsam}, which has improved robustness on the degraded dataset. 
3) Additionally, we compare with two-stage methods, \textit{i.e.}, reconstructing images first with image reconstruction (IR) networks and passing the restored clear images to the SAM and SAM2. We use two state-of-the-art IR networks for comparison: PromptIR \cite{promptir}, and diffusion-based DiffBIR \cite{diffbir}. 
4) Furthermore, we extended our experiments to include fine-tuning of PromptIR and DiffBIR using our degraded-clear image pairs to more comprehensively validate the superiority of our approach, referred to as PromptIR-FT and DiffBIR-FT. 
We employ three metrics to assess our model's performance, including  Intersection over Union (IoU), Dice Coefficient (Dice), and Pixel Accuracy (PA).    

\begin{table}[t] \footnotesize
\caption{Performance comparison on the test set of Thinobject-5K \cite{thin} and LVIS \cite{lvis} datasets (In-distribution evaluation) with different levels of degradation. From LQ-1 to LQ-3, the degree of degradation increases progressively. We report IoU and Dice for comparison. Our GleSAM++ and GleSAM2++ consistently outperform other competitors, especially on the most challenging LQ-3 version.
The words with boldface indicate the best results, and those underlined indicate the second-best results.}
\centering
\renewcommand\arraystretch{1.2}
\setlength{\tabcolsep}{0.7 mm}{
\begin{tabular}{ccccccc|cccccc}\toprule[0.8pt]
     \multirow{3}{*}{{Method}} & \multicolumn{6}{c|}{ThinObject-5K} & \multicolumn{6}{c}{LVIS} \\
     & \multicolumn{2}{c}{{LQ-3}} & \multicolumn{2}{c}{{LQ-2}} &\multicolumn{2}{c|}{{LQ-1}} & \multicolumn{2}{c}{{LQ-3}} & \multicolumn{2}{c}{{LQ-2}} &\multicolumn{2}{c}{{LQ-1}} \\
\cmidrule(r){2-3}  \cmidrule(r){4-5} \cmidrule(r){6-7} \cmidrule(r){8-9}  \cmidrule(r){10-11} \cmidrule(r){12-13}
                               & IoU & Dice & IoU & Dice 
                            & IoU & Dice & IoU & Dice & IoU & Dice 
                            & IoU & Dice \\
\toprule[0.5pt]
{SAM}   & 0.633 & 0.733  & 0.700 &0.789 & 0.736 & 0.820 & 0.406  & 0.504 &0.474 & 0.572 & 0.528 & 0.626 \\
{RobustSAM}  & 0.695 & 0.791 & 0.754  & 0.838 & 0.778 & {0.855} & 0.455  & 0.574 & 0.492 & 0.606 & 0.526   & 0.636 \\
{PromptIR-SAM} & 0.630 & 0.726 &0.705 & 0.790 & 0.737 & 0.817 & 0.401 & 0.489 & 0.469 & 0.557 & 0.522 & 0.610 \\
{PromptIR-FT-SAM} & 0.636 & 0.736 & 0.710 & 0.804 & 0.740 & 0.820 & 0.409 & 0.495 & 0.476 & 0.560 & 0.537 & 0.617  \\
DiffBIR-SAM  & {0.684} & {0.780} & {0.762} & {0.839} & {0.778} & 0.865  & {0.526} & {0.632} & {0.574} & {0.691} & {0.591} & {0.708} \\
{DiffBIR-FT-SAM} & {0.699} & {0.802} & {0.775} & {0.847} & {0.790} & 0.870  & {0.539} & {0.646} & {0.579} & \underline{0.708} & {0.601} & {0.716} \\
{GleSAM} & \underline{0.737} & \underline{0.825}  & \underline{0.782}  & \underline{0.858} & \underline{0.799} & \underline{0.871} & \underline{0.548} & \underline{0.675} & \underline{0.590} & 0.704 & \underline{0.602} & \underline{0.718} \\
{{GleSAM++ (Ours)}} & \textbf{0.770} & \textbf{0.848} & \textbf{0.805} & \textbf{0.873} & \textbf{0.826} & \textbf{0.891} & \textbf{0.553} & \textbf{0.682} & \textbf{0.596} & \textbf{0.719} & \textbf{0.612} & \textbf{0.731} \\
\midrule[0.5pt]
{SAM2}  &0.733 & 0.812 &0.766 & 0.839 & 0.782 &  0.852  & 0.520 &0.625 & 0.561 &0.663 & 0.595 & 0.694  \\
{PromptIR-SAM2} & 0.730 & 0.814 & 0.771 &0.844 & 0.781 & 0.849 & 0.516 & 0.620 & 0.558 & 0.663 & 0.589 & 0.690 \\
{PromptIR-FT-SAM2} & 0.737 & 0.819 & 0.791 &0.856 & 0.795 & 0.852 & 0.521 & 0.623 & 0.562 & 0.671 & 0.601 & 0.702  \\
DiffBIR-SAM2 & {0.735}  & {0.812} & {0.783} & {0.851} & {0.797} & {0.860} & {0.560} & {0.667} & {0.597} & {0.700} & {0.615} &  {0.724} \\
{DiffBIR-FT-SAM2} & 0.740 & 0.821 & 0.791 & 0.857 & 0.798 & 0.865 & 0.562 & 0.671 & 0.597 & 0.703 & 0.618 & 0.718
\\
{GleSAM2} & \underline{0.752} & \underline{0.842} & \underline{0.802} & \textbf{0.873} & \underline{0.812} & \underline{0.879} & \underline{0.572} & \underline{0.680} & \underline{0.601} & \underline{0.711}  & \underline{0.634} & \underline{0.736} \\
{GleSAM2++ (Ours)} & \textbf{0.776} & \textbf{0.850} & \textbf{0.807} & \underline{0.871} & \textbf{0.834} & \textbf{0.894} & \textbf{0.584} & \textbf{0.709} & \textbf{0.622} & \textbf{0.740} & \textbf{0.643} & \textbf{0.756}  \\
\toprule[0.8pt]
\end{tabular}}
\label{tab:thin}
\vspace{-5pt}
\end{table}

\begin{table}[h] \footnotesize
\caption{Performance comparison on the ECSSD \cite{ecssd} and COCO \cite{coco} datasets (cross-dataset evaluation) with different levels of degradation. These results indicate that GleSAM++/GleSAM2++ possess significant robustness in cross-dataset segmentation across different levels of degradation.}
\centering
\renewcommand\arraystretch{1.2}
\setlength{\tabcolsep}{0.6 mm}{
\begin{tabular}{ccccccc|cccccc}\toprule[0.8pt]
     \multirow{3}{*}{{Method}} & \multicolumn{6}{c|}{{ECSSD}} & \multicolumn{6}{c}{{COCO}} \\ 
     & \multicolumn{2}{c}{{LQ-3}} & \multicolumn{2}{c}{{LQ-2}} & \multicolumn{2}{c|}{{LQ-1}} & \multicolumn{2}{c}{{LQ-3}} & \multicolumn{2}{c}{{LQ-2}}  & \multicolumn{2}{c}{{LQ-1}} \\ 
     \cmidrule(r){2-3}  \cmidrule(r){4-5} \cmidrule(r){6-7}  \cmidrule(r){8-9}  \cmidrule(r){10-11} \cmidrule(r){12-13}
                               & IoU  & Dice & IoU & Dice & IoU  & Dice & IoU & Dice & IoU  & Dice & IoU & Dice \\
\toprule[0.5pt]
{SAM}   &0.525 &0.643  & 0.608 & 0.715 & 0.676 & 0.770 & 0.409 & 0.507 & 0.487 & 0.583 & 0.534 & 0.630 \\
{RobustSAM}  & 0.655 & {0.769} & 0.722 & 0.821  & {0.762} & {0.849} & 0.456  & 0.574  & 0.501 &0.614 &0.532 & 0.642 \\
{PromptIR-SAM} & 0.526 & 0.644 & 0.613 &0.710 & 0.672 &0.768  & 0.401  & 0.498 & 0.467 & 0.574 & 0.528 & 0.620 \\
{PromptIR-FT-SAM} & 0.536 & 0.653 & 0.621 &0.717 & 0.685 & 0.773 & 0.416 & 0.503 & 0.475 & 0.582 & 0.540 & 0.635 \\
{DiffBIR-SAM} &{0.655} & 0.757 & {0.732} & {0.814} & 0.762 & 0.843 & {0.515} & {0.621} & {0.563} & {0.683} & {0.604} & {0.702}\\
{DiffBIR-FT-SAM} & {0.668} &0.772 &{0.749} &{0.832} & 0.777 &0.858 &{0.520} &{0.633} &{0.568} & {0.688} & {0.606} & {0.709} \\
{GleSAM} & \underline{0.701}  & \underline{0.804} &  \underline{0.762} & \underline{0.849} & \underline{0.791} & \underline{0.869} & \underline{0.547} & \underline{0.670} & \underline{0.580}& \underline{0.695} & \underline{0.610}& \underline{0.721}\\
{GleSAM++ (Ours)} & \textbf{0.741} & \textbf{0.832} & \textbf{0.797} & \textbf{0.872} & \textbf{0.816} & \textbf{0.886} & \textbf{0.569} & \textbf{0.697} & \textbf{0.603} & \textbf{0.724} & \textbf{0.621} & \textbf{0.738} \\
\midrule[0.5pt]
{SAM2} &0.657 & 0.757 & 0.736 &0.821 & 0.772 & 0.848 & 0.519 &0.625 & 0.570 & 0.670 & 0.606 &0.702 \\
{PromptIR-SAM2} & 0.648  & 0.757  & 0.720  &0.813 & 0.757 & 0.837 & 0.509 &0.617  & 0.559 &0.663 & 0.593 & 0.693\\
{PromptIR-FT-SAM}  & 0.653  & 0.758  & 0.723  &0.818 & 0.762 & 0.840 & 0.514 &0.619  & 0.565 & 0.668 & 0.598 & 0.698 \\
{DiffBIR-SAM2} & {0.680} & {0.782} & {0.754} & {0.829} & {0.776} & {0.846}  & {0.563}  & {0.670} & {0.604} & {0.713} & {0.621} & {0.728} \\
{DiffBIR-FT-SAM} & 0.684 & 0.785 & \underline{0.757} & 0.833 & 0.780 & 0.851 & 0.569 & 0.674 & 0.610 & 0.718 & 0.623 & 0.730 \\
{GleSAM2} & \underline{0.694} & \underline{0.794} & \underline{0.757} & \underline{0.843} & \underline{0.789} & \underline{0.864} & \underline{0.572} & \underline{0.687} & \underline{0.617} & \underline{0.725} & \underline{0.634} & \underline{0.737} \\
{GleSAM2++ (Ours)} & \textbf{0.705} & \textbf{0.802} & \textbf{0.773} & \textbf{0.854} &  \textbf{0.795} & \textbf{0.867} & \textbf{0.595} & \textbf{0.719} & \textbf{0.632} & \textbf{0.748} & \textbf{0.647} & \textbf{0.760} \\
\toprule[0.8pt]
\end{tabular}}
\label{tab:unseen-sets}
\vspace{-5pt}
\end{table}

\begin{table}[t] \small
\caption{Zero-shot performance comparison on Robustseg-style \cite{robustsam} and real-world degradations. Note that our methods are not trained on such degradations. Performance is tested on the  ECSSD, COCO, and BDD-10K datasets. The superior performance of our method demonstrates robustness against various degradations.}
\centering
\renewcommand\arraystretch{1.1}
\setlength{\tabcolsep}{2.5 mm}{
\begin{tabular}{ccccccc}\toprule[0.8pt]
     \multirow{3}{*}{{Method}} & \multicolumn{4}{c}{Robustseg-Style Degradation} & \multicolumn{2}{c}{Real-World Degradations}  \\ \cmidrule(r){2-5} \cmidrule(r){6-7}
     &\multicolumn{2}{c}{{ECSSD}} &\multicolumn{2}{c}{{COCO}} & \multicolumn{2}{c}{BDD-10K} \\
\cmidrule(r){2-3} \cmidrule(r){4-5}  \cmidrule(r){6-7}
                         & IoU & Dice & IoU & Dice  & IoU & Dice \\
\toprule[0.5pt]
SAM & 0.794 & 0.863 & 0.659 & 0.751 & 0.865 &  0.922  \\
RobustSAM & 0.858 & 0.905 & 0.624 & 0.728 &0.864 & 0.918  \\
{GleSAM}  & 0.850 & 0.907 & 0.663 & 0.772  &{0.873} & {0.924}\\
{GleSAM++ (Ours)} & \textbf{0.861} & \textbf{0.913} & \textbf{0.668} & \textbf{0.775} & \textbf{0.879} & \textbf{0.930}  \\
\hline
{SAM2}  & 0.825 & 0.884 & 0.685 & 0.775 & 0.889 & 0.937  \\
{GleSAM2}  &{0.853}  & {0.898} &{ 0.685 }& {0.779} &{0.900} & {0.940} \\
GleSAM2++ (Ours) &\textbf{0.857}  & \textbf{0.909} &\textbf{0.703}& \textbf{0.803} & \textbf{0.903} & \textbf{0.947}  \\
\toprule[0.8pt]
\end{tabular}}
\label{tab:bdd}
\vspace{-5pt}
\end{table}

\begin{figure*}[t]
  \centering
  \includegraphics[width=0.97\linewidth]{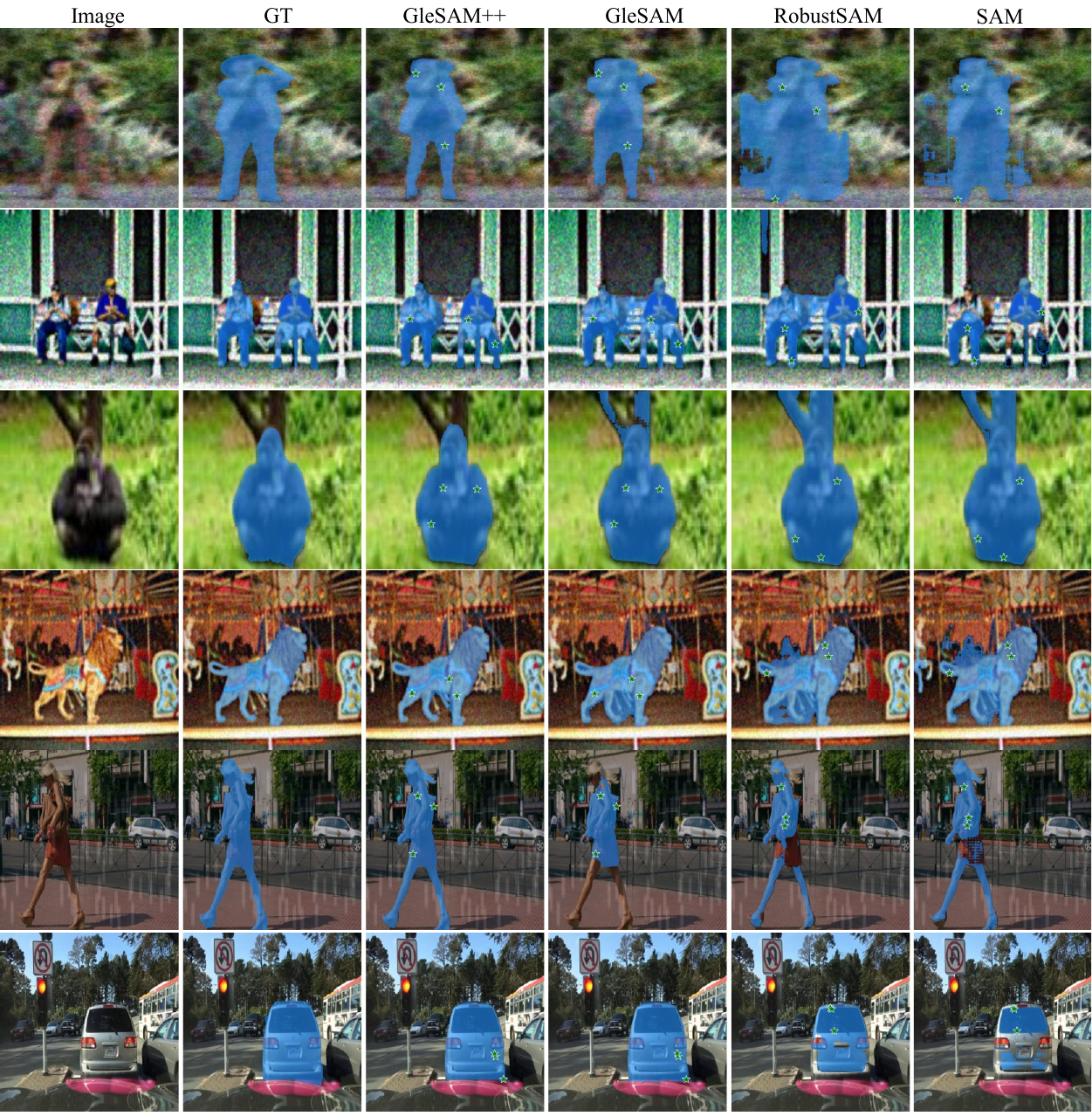}
   \caption{Visual comparisons on the ECSSD, Robust-Seg, and BDD-10K datasets. The results demonstrate the superior generalization capability of GleSAM++ to handle unseen degradations
not included in the training set.
}
   \label{fig:results_vis}
\end{figure*}

\subsection{Performance Comparisons}
In this experiment, we evaluate the performance on the test set of our LQSeg, including in-distribution (Tab. \ref{tab:thin}) and cross-dataset (Tab. \ref{tab:unseen-sets}) evaluations.


\subsubsection{Main results}
In Tab. \ref{tab:thin}, we evaluate the performance of our GleSAM++ and GleSAM2++ on two in-distribution datasets: ThinObject-5K \cite{thin} and LVIS \cite{lvis}. Each dataset contains three levels of degradation. 
Our GleSAM++/GleSAM2++ achieve the best performance across all three degradation levels on both datasets. Specifically, on ThinObject-5K (LQ-3), GleSAM++ improves IoU by +13\% and Dice by +11\% over the baseline (SAM).
The performance gap widens as degradation severity increases (from LQ-1 to LQ-3), indicating that our degradation-aware generative adaptation substantially enhances model robustness.
For GleSAM2++, the advantage is also evident: it surpasses SAM2 by +6\% IoU and +8\% Dice under the LVIS dataset (LQ-3), confirming its capability in maintaining segmentation accuracy under complex degradations. Furthermore, GleSAM++ and GleSAM2++ also outperform their predecessors, \textit{i.e.}, GleSAM and GleSAM2, demonstrating the improved robustness and generalization.

Tab. \ref{tab:unseen-sets} presents the segmentation performance of GleSAM++/GleSAM2++ on two cross-dataset evaluations: ECSSD \cite{ecssd} and COCO \cite{coco}. GleSAM++ and GleSAM2++ also consistently outperform other baseline methods, particularly on the most challenging LQ-3 version, underscoring their strong generalization capabilities and potential for real-world applications.

\subsubsection{Validation with other degradations}
To validate the model's generalization on other unseen degradations, we then evaluated GleSAM++/GleSAM2++ on unseen degradation types, including both \textbf{RobustSeg-style} synthetic degradations \cite{robustsam} and \textbf{real-world} degradations \cite{bdd100k}. The quantitative results are summarized in Tab.~\ref{tab:bdd}.
These degradations were not used during training. In this evaluation, we compare our methods with SAM, SAM2, RobustSAM, and the previous GleSAM and GleSAM2 to assess their relative performance. 
For {RobustSeg-style degradations}, our methods consistently outperform SAM and SAM2, and even surpass RobustSAM, which is explicitly trained on such degradations. This indicates that our method effectively improves the intrinsic robustness of the segmentation process, even without degradation-specific training. We further extend our evaluation to the {BDD-10K} dataset~\cite{bdd100k}, which involves complex real-world degradations such as adverse weather (\eg, rain, fog) and dynamic illumination conditions. These uncontrolled scenarios pose significant challenges for segmentation generalization. As shown in Tab.~\ref{tab:bdd}, GleSAM++ and GleSAM2++ maintain superior performance across all metrics, demonstrating remarkable adaptability to natural degradations.



\subsubsection{Validation of general segmentation ability}
\label{sec:generalization_clear}
\begin{wrapfigure}{r}{0.4\textwidth} 
    \centering
        \includegraphics[width=\linewidth]{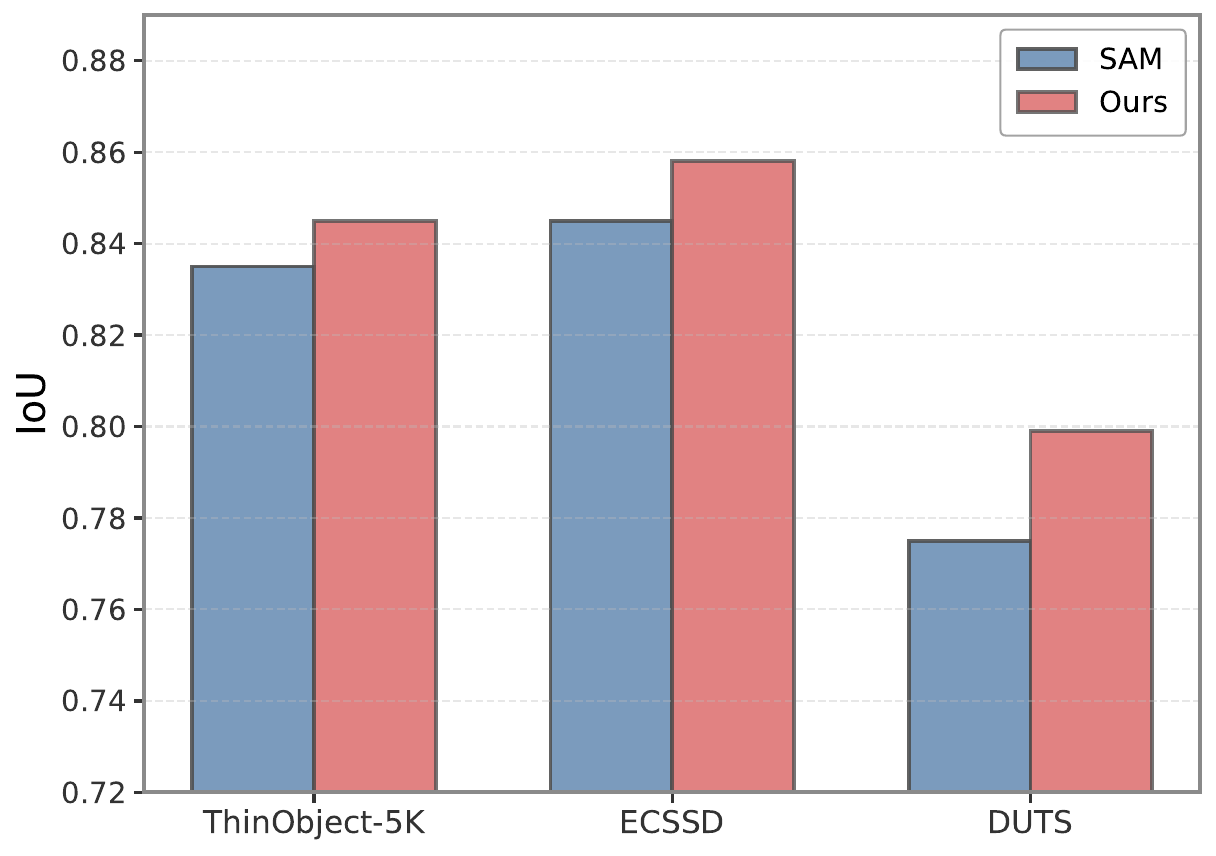}
        \caption{{Comparison of SAM and GleSAM++ on clear-image benchmarks.}}
        \label{fig:clear_generalization}
    \vspace{-20pt}
\end{wrapfigure}
In this part, we further examine whether GleSAM++ preserves the general segmentation ability of SAM after domain-specific fine-tuning. 
The results on clear-image benchmarks are shown in Fig.~\ref{fig:clear_generalization}. We conduct the evaluation on ThinObject-5K and ECSSD, and further include DUTS \cite{duts} as an additional unseen benchmark to assess cross-dataset generalization.
It can be observed that, even after domain-specific finetuning, GleSAM++ still maintains strong general segmentation performance, and consistently outperforms the original SAM across all evaluated datasets.

{
This observation indicates that the proposed method does not sacrifice the original general-purpose capability of the foundation model. We attribute this property to two factors. 1) GleSAM++ performs generative feature enhancement rather than replacing the original segmentation representation, and thus tends to preserve high-quality features when the input is already clear. 2) The SAM image encoder remains frozen during training, which helps retain the general visual representation learned from large-scale pre-training. As a result, the proposed fine-tuning strategy improves robustness to degraded inputs while maintaining, and in some cases even enhancing, the general segmentation ability on clear images.}

\vspace{10pt}
\subsubsection{Visualizations}
Fig. \ref{fig:results_vis} showcases the qualitative comparisons of SAM, RobustSAM, GleSAM, and GleSAM++ under various degradations. 
Due to the challenging degradations, previous SAM and the enhanced RobustSAM struggle to segment these objects accurately, resulting in serious detail missing and erroneous background prediction, showing their limitations.
GleSAM achieves noticeably better segmentation results by introducing generative enhancement. However, its fixed-strength reconstruction limits adaptability across varying degradation severities.
In contrast, GleSAM++ effectively recovers finer details and achieves more precise segmentation results. 
These visual results highlight the robustness of our method in processing low-quality images.

\begin{table*}[t] \footnotesize
\caption{Ablation study of each component in the proposed method, evaluated on the ECSSD and COCO datasets. Each additional component positively affects the performance, demonstrating the effectiveness of the proposed methods.}
\centering
\renewcommand\arraystretch{1.25}
\setlength{\abovecaptionskip}{0pt}%
\setlength{\belowcaptionskip}{5pt}%
\setlength{\tabcolsep}{0.26 mm}{
\begin{tabular}{lcccccc|cccccc}\toprule[0.8pt]
     \multirow{3}{*}{{Method}} & \multicolumn{6}{c|}{{ECSSD}} & \multicolumn{6}{c}{{COCO}} \\ 
     & \multicolumn{2}{c}{{LQ-3}} & \multicolumn{2}{c}{{LQ-2}} & \multicolumn{2}{c|}{{LQ-1}} & \multicolumn{2}{c}{{LQ-3}} & \multicolumn{2}{c}{{LQ-2}}  & \multicolumn{2}{c}{{LQ-1}} \\ 
     \cmidrule(r){2-3}  \cmidrule(r){4-5} \cmidrule(r){6-7}  \cmidrule(r){8-9}  \cmidrule(r){10-11} \cmidrule(r){12-13}
                               & IoU  & Dice & IoU & Dice & IoU  & Dice & IoU & Dice & IoU  & Dice & IoU & Dice \\
\toprule[0.5pt]
Baseline &0.525 &0.643  & 0.608 & 0.715 & 0.676 & 0.770 & 0.409 & 0.507 & 0.487 & 0.583 & 0.534 & 0.630 \\
+ Gle \& CRE  & 0.573 & 0.683 & 0.651 & 0.754 & 0.699 & 0.801 & 0.426 & 0.528 & 0.502 & 0.602 & 0.549 & 0.650 \\
+ Gle \& CRE \& FDA & 0.613 & 0.728 & 0.684 & 0.785 & 0.723 &0.826 & 0.444 & 0.545 & 0.518 & 0.617 & 0.565 & 0.662   \\
+ Gle \& CRE \& FDA \& DAE  &\textbf{0.686} & \textbf{0.783} & \textbf{0.751} & \textbf{0.835} & \textbf{0.785} & \textbf{0.860} & \textbf{0.466} & \textbf{0.568} & \textbf{0.535} & \textbf{0.636} & \textbf{0.575} & \textbf{0.675} \\
\toprule[0.8pt]
\end{tabular}}
\label{tab:ablation}
\vspace{-5pt}
\end{table*}

\begin{table*}[t] \footnotesize
\caption{{Effect of Fine-tuning SAM. The performance is evaluated on the ECSSD dataset. ``FT-T'' and ``FT-D'' indicate fine-tuning the SAM's mask token and decoder, respectively. ``Clear'' indicates the results on the original clear images.}}
\centering
\renewcommand\arraystretch{1.25}
\setlength{\tabcolsep}{0.48 mm}{
\begin{tabular}{ccccccccccccc}\toprule[0.8pt]
     \multirow{2}{*}{{Method}} & \multicolumn{2}{c}{{LQ3}} & \multicolumn{2}{c}{{LQ2}} & \multicolumn{2}{c}{{LQ1}} & \multicolumn{2}{c}{{Robust-style}} &\multicolumn{2}{c}{{Clear}} & \multicolumn{2}{c}{{Average}}\\
\cmidrule(r){2-3}  \cmidrule(r){4-5} \cmidrule(r){6-7} \cmidrule(r){8-9} \cmidrule(r){10-11}
                               & IoU & Dice  & IoU & Dice  & IoU & Dice & IoU & Dice & IoU & Dice & IoU & Dice 
                             \\
\toprule[0.5pt]
\multicolumn{5}{l}{\textbf{\textit{w/o Fine-tuning SAM:}}} \\
SAM &0.525 &0.643  & 0.608 & 0.715 & 0.676 & 0.770 & 0.794 & 0.863 & 0.820 & 0.879 & 0.571 & 0.645 \\
GleSAM++ (Ours) & 0.686 &0.783 &0.751 &0.835 & 0.785 & 0.860 & 0.816 & 0.881 &0.843 &0.897  & 0.647 & 0.709  \\
\midrule[0.5pt]
\multicolumn{5}{l}{\textbf{\textit{Fine-tuning SAM: }}} \\
SAM-FT-T   & 0.642 &0.758 &0.710 &0.809 & 0.757 & 0.845 & 0.832 & 0.888 & 0.852 & 0.904 & 0.631 & 0.701 \\
SAM-FT-D  & 0.673 &0.783 &0.733 &0.826 & 0.776 & 0.860 & 0.838 & 0.894 & 0.865 & 0.907 & 0.648 & 0.712 \\
GleSAM++-FT-T (Ours) &0.729 &0.821 & 0.783 & 0.864 & 0.808 & 0.879 & 0.856 & 0.910 & 0.881  & 0.925 & 0.676  & 0.733 \\
GleSAM++-FT-D (Ours) & \textbf{0.741} & \textbf{0.832} & \textbf{0.797} & \textbf{0.872} & \textbf{0.816} & \textbf{0.885}  & \textbf{0.869} & \textbf{0.930} & \textbf{0.894} & \textbf{0.934} &  \textbf{0.686} & \textbf{0.741} \\
\toprule[0.8pt]
\end{tabular}}
\label{tab:ablation-ft}
\end{table*}



\subsection{Ablation study}
\label{sec:ablation}

\begin{wrapfigure}{r}{0.55\textwidth} 
    \centering
    \vspace{-20pt}
        \includegraphics[width=\linewidth]{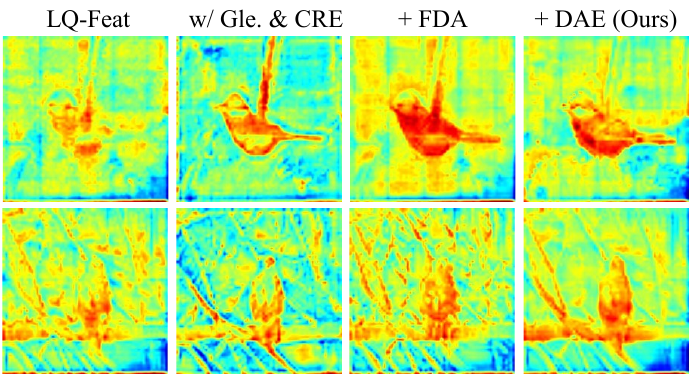}
        \caption{Qualitative visualization of the enhanced latent features. The clearest feature is obtained when combining all modules, \textit{i.e.}, Gle+CRE+FDA+DAE.} 
        \vspace{-20pt}
    \label{fig:ablation-vis}
\end{wrapfigure}
As summarized in Tab. \ref{tab:ablation}, we conduct an ablation study to assess the individual contributions of the proposed modules in GleSAM++, namely the Feature Distribution Alignment (FDA), Channel Replication and Expansion (CRE), and Degradation-aware Adaptive Enhancement (DAE). 
Here, “Gle” denotes the base framework equipped with generative latent space enhancement. To ensure a fair comparison of reconstruction capability, all experiments are performed by fine-tuning only the U-Net backbone.

\subsubsection{Effect of each component}
\label{sec: ablation-modules}

The quantitative results demonstrate that each additional module consistently improves segmentation accuracy across all degradation levels and datasets. Notably, integrating FDA and CRE significantly bridges the gap between the pre-trained latent diffusion model and SAM, while introducing DAE further boosts performance by adaptively regulating the denoising strength. 
The highest IoU and Dice scores are achieved when combining all modules.
Qualitative results in Fig.~\ref{fig:ablation-vis} further validate these findings: the reconstructed latent features become progressively clearer and more structured as additional modules are incorporated.
Moreover, as shown in Tab. \ref{tab:ablation-ft}, even without fine-tuning the decoder, the proposed method improves performance on low-quality images by approximately 7 percentage points, while maintaining robustness on high-quality inputs. These results underscore the effectiveness and generalization capability of our generative enhancement framework.


\subsubsection{Effect of fine-tuning SAM} 
Tab. \ref{tab:ablation-ft} investigates the influence of decoder fine-tuning on segmentation performance. Two configurations are examined: fine-tuning SAM’s entire decoder and fine-tuning only the output mask token. Results indicate that directly fine-tuning SAM on degraded images improves performance under low-quality conditions, highlighting the role of data adaptation. However, the performance on degraded inputs still lags significantly behind that on clear images, with nearly a \textbf{20-point drop} in IoU. In contrast, our GleSAM++ substantially mitigates this performance gap and further boosts accuracy on both degraded and clear images when fine-tuning is applied, demonstrating robust adaptation across arbitrary image qualities.

\begin{figure*}[t]
  \centering
 \includegraphics[width=\linewidth]{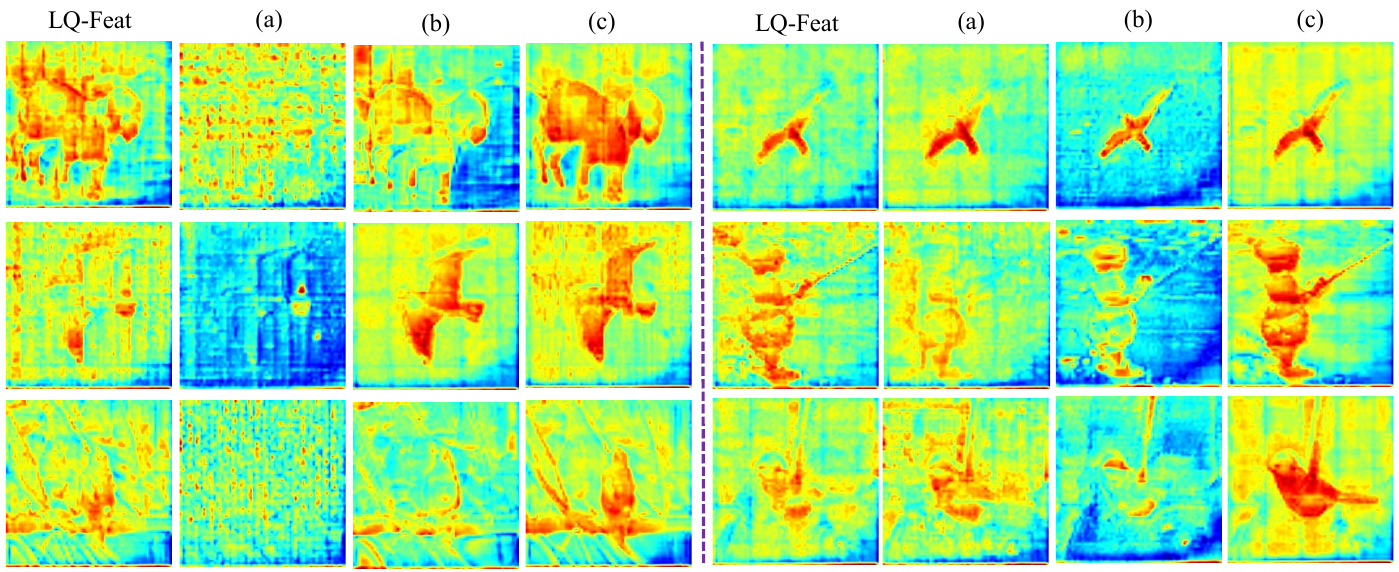}
    \caption{{Qualitative visualization of the enhanced latent features generated by different channel expansion methods. Our CRE method (c) produces more salient features.}}
   \label{fig:ablation-vis-cre}
   \vspace{-10pt}
\end{figure*}

\subsubsection{Analysis of channel expansion methods}
\label{sec: ablation-cre}
To address the channel mismatch between SAM and the pre-trained U-Net, we explored three strategies: (a) using two simple convolutional layers to reduce and expand the channels of segmentation features as needed, (b) fine-tuning new head and tail layers from scratch, and (c) our proposed Channel Replication and Expansion method. 
The results are tested without finetuning the decoder.
As shown in Tab. \ref{tab:ablation-REI} and Fig. \ref{fig:ablation-vis-cre}, strategies (a) and (b) achieve limited improvements, likely because the new layers couldn’t leverage the pre-trained knowledge. 
In contrast, our CRE method resolves the channel inconsistency by replicating pre-trained weights and refining them via LoRA, leading to consistently superior performance across all metrics.

\begin{table}[t] \small
\vspace{-10pt}
\caption{Analysis of the proposed CRE. It significantly outperforms alternative approaches, achieving higher scores.}
\centering
\renewcommand\arraystretch{1.25}
\setlength{\tabcolsep}{4 mm}{
\begin{tabular}{c|ccc}\toprule[0.8pt]
     {Method} & IoU & Dice  & PA \\
\toprule[0.5pt]
(a) Learnable projection layers & 0.449 &  0.586 &  0.620 \\
(b) Training head and tail layers from scrith & 0.617 &0.705 & 0.783 \\
(c) Replicate and Expansion (Ours) & \textbf{0.651} & \textbf{0.754} & \textbf{0.844} \\
\toprule[0.8pt]
\end{tabular}}
\label{tab:ablation-REI}
\vspace{-20pt}
\end{table}


\begin{table}[t]
\centering
\begin{minipage}{0.7 \linewidth}
\centering
\footnotesize
\caption{Analysis of the proposed DAE. Our method achieves superior performance to directly using $s$. }
\renewcommand\arraystretch{1.25}
\setlength{\tabcolsep}{2. mm}{
\begin{tabular}{c|ccc}\toprule[0.8pt]
     {Method} & IoU & Dice  & PA \\
\toprule[0.5pt]
(a) Directly use $s$ & 0.754 & 0.838 & 0.911 \\
(b) Ours  & \textbf{0.797} & \textbf{0.872} & \textbf{0.956} \\
\toprule[0.8pt]
\end{tabular}}
\label{tab:ablation-dae}
\end{minipage}
\hspace{0.01\linewidth} 
\begin{minipage} {0.3 \linewidth}
\centering
\footnotesize
\caption{{Analysis of LoRA ranks in U-Net.}}
\renewcommand\arraystretch{1.25}
\setlength{\tabcolsep}{1. mm}{
\begin{tabular}{ccccc}\toprule[0.8pt]
    {Rank} & IoU & Dice & PA & Params\\
\toprule[0.5pt]
4  & 0.780 & 0.861 & 0.944 & 16.25M \\
8  & \textbf{0.784} & \textbf{0.864} & \textbf{0.946}  & 32.49M \\
16  & 0.733 & 0.826 & 0.928 & 64.99M \\
\toprule[0.8pt]
\end{tabular}}
\label{tab:rank}
\end{minipage}
\vspace{-20pt}
\end{table}

\subsubsection{Analysis of DAE method}
\begin{wrapfigure}{r}{0.45\textwidth} 
    \centering
        \includegraphics[width=\linewidth]{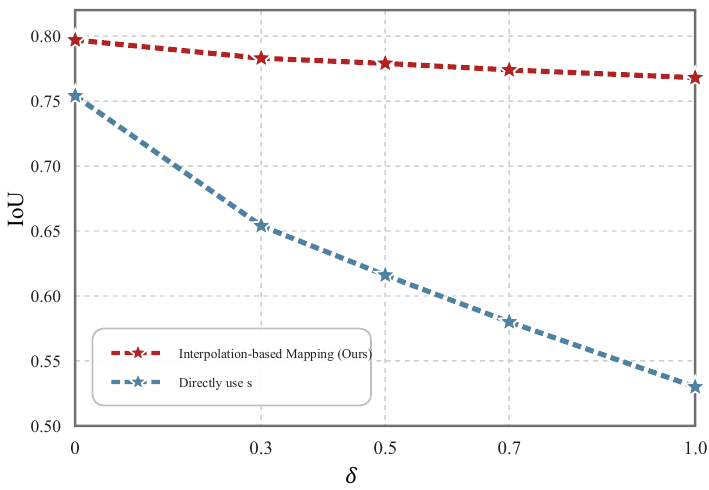}
        \caption{{Sensitivity analysis of the DAE mechanism under different perturbation magnitudes on the predicted degradation score.}}
        \label{fig:sensity-test}
    \vspace{-25pt}
\end{wrapfigure}
Table~\ref{tab:ablation-dae} compares two generative enhancement strategies conditioned on the predicted degradation score: (a) directly using $s$ to control the denoising intensity, and (b) our proposed degradation-aware interpolation scheme. 
Although $s$ is a dynamic signal, we observe that directly substituting it into the diffusion parameters leads to training instability and suboptimal reconstruction quality. 
This issue arises because the predicted $s$ may contain noise, causing a mismatch with the noise schedule learned by the diffusion model. 
To address this, we perform linear interpolation between predefined minimum and maximum noise levels, constraining the adaptive noise ratio within an effective range and regularizing the mapping from $s$ to the diffusion parameters. This design preserves the mathematical consistency of the diffusion process, stabilizes training, and yields significantly better and more reliable reconstruction results.

{To further evaluate the robustness of DAE under imperfect score estimation, we perform an additional sensitivity analysis by perturbing the predicted degradation score with different magnitudes. As shown in Fig. \ref{fig:sensity-test}, the interpolation-based mapping exhibits only a mild and gradual performance drop as the perturbation increases, whereas directly using $s$ results in a much steeper degradation trend. This observation indicates that the proposed mapping strategy effectively regularizes the influence of score errors and prevents them from being excessively amplified during denoising. Therefore, beyond improving the average reconstruction quality, the proposed DAE design also provides stronger robustness and bounded sensitivity to inaccurate degradation prediction.}

\subsubsection{Analysis of denoising steps}
We further conduct a systematic ablation study by varying the number of denoising steps from 1 to 5 under three degradation levels. We claim that one-step denoising provides the most favorable balance between segmentation performance and computational efficiency. As reported in Tab. \ref{tab:ablation-denoising-step}, the best segmentation performance is achieved within a very small number of denoising steps. Importantly, the performance does not continue to improve as the number of steps increases. Instead, it gradually declines after the third step. This trend suggests that a limited amount of iterative refinement is beneficial for restoring degraded semantic features, whereas excessive denoising tends to cause over-correction or over-smoothing, thereby weakening boundary details and discriminative cues that are critical for accurate segmentation.

We also report the inference efficiency under different denoising steps. Although increasing the denoising process from 1 step to 3 steps brings marginal performance gains, the additional computational cost and latency are disproportionately larger. Therefore, 1-step denoising offers the most favorable trade-off between segmentation performance and inference efficiency. This observation further supports our design choice of single-step denoising as a practical and efficient default setting.

\begin{table*}[t] \footnotesize
\caption{{Ablation study on the number of denoising steps under three degradation levels. Single-step denoising provides the best accuracy-efficiency trade-off.}}
\centering
\renewcommand\arraystretch{1.25}
\setlength{\tabcolsep}{2.5 mm}{
\begin{tabular}{cccccccccc}\toprule[0.8pt]
     \multirow{2}{*}{{Step}} & \multicolumn{2}{c}{{LQ3}} & \multicolumn{2}{c}{{LQ2}} & \multicolumn{2}{c}{{LQ1}} & \multicolumn{2}{c}{{Average}} & {Latency} \\
\cmidrule(r){2-3}  \cmidrule(r){4-5} \cmidrule(r){6-7} \cmidrule(r){8-9} 
                               & IoU & Dice  & IoU & Dice  & IoU & Dice & IoU & Dice (s) 
                             \\
\toprule[0.5pt]
Baseline & 0.525	&0.643	&0.608	&0.715	&0.676	&0.770	&0.603	&0.709  & 0.32 \\
1 &0.730    &0.824	&0.773	&0.855	&0.811	&0.882 &0.771 & 0.854  & 0.38$_{\uparrow0.06}$ \\
2 &0.747	&0.836	&0.788	&0.865	&0.822	&0.889  &0.786 & 0.863 & 0.51$_{\uparrow0.19}$ \\
3 &0.748	&0.836	&0.790	&0.867	&0.821	&0.888  &0.786 & 0.864 & 0.56$_{\uparrow0.24}$ \\
4 &0.739    &0.829	&0.783	&0.861	&0.813	&0.883  &0.778 & 0.858 & 0.66$_{\uparrow0.34}$ \\
5 &0.729	&0.821	&0.770	&0.853	&0.802	&0.876  & 0.767 & 0.850  &0.81$_{\uparrow0.49}$ \\
\toprule[0.8pt]
\end{tabular}}
\label{tab:ablation-denoising-step}
\end{table*}

\begin{figure}[t]
  \centering
   \includegraphics[width=0.85 \linewidth]{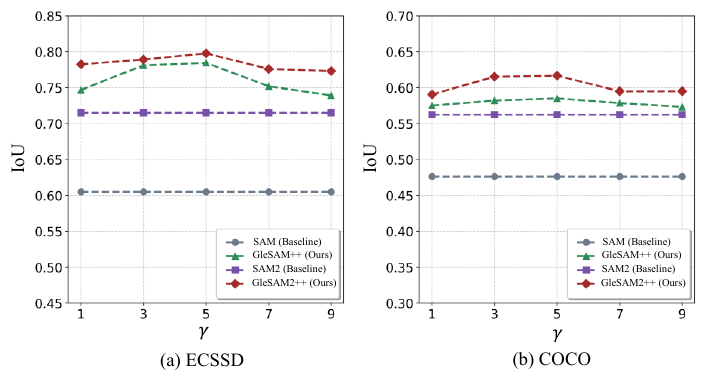}
   \caption{Ablation study of adaption weight $\gamma$. }
   \label{fig:ablation-k}
   \vspace{-10pt}
\end{figure}

\subsubsection{Analysis of LoRA ranks and hyperparameter $\gamma$}
In this part, we analyzed in detail the effects of two hyperparameters: LoRA ranks and $\gamma$. We integrate learnable LoRA layers into the pre-trained denoising U-Net to enable efficient fine-tuning. To assess the influence of different LoRA ranks on segmentation performance, we conduct experiments with ranks of 4, 8, and 16, as reported in Tab.~\ref{tab:rank}. The results show that a rank of 8 achieves the best overall performance while keeping the number of learnable parameters at a moderate level, demonstrating a favorable trade-off between accuracy and efficiency.


We use an adaptation weight $\gamma$ to align the distribution of the latent space between LDM and SAM. 
To determine the optimal value of $\gamma$, we empirically test five different values on the ECSSD and COCO datasets. The results, shown in Fig. \ref{fig:ablation-k}, suggest that $\gamma = 5$ is the most effective setting, providing strong generalization across all models and datasets. Therefore, we adopt $\gamma = 5$ as the default value in our experiments.

\subsubsection{Analysis of two-stage learning}
We further analyze the effect of the proposed two-stage learning strategy. Compared with single-stage joint optimization, this design reduces the interference between feature restoration and segmentation learning, leading to more stable and effective training. 
As shown in Tab. \ref{tab:ablation-twostage-training}, we report the averaged results on the four datasets and find that two-stage learning consistently improves segmentation performance across all degradation levels. 
Importantly, even when the number of training iterations for single-stage training is substantially increased, its performance still remains inferior to that of the proposed two-stage strategy. This suggests that the benefit of the two-stage design does not simply arise from longer training, but from a more effective optimization process that better decouples feature restoration and segmentation learning.

\begin{table*}[t] \footnotesize
\caption{{Effect of the two-stage learning strategy, which leads to more stable optimization and stronger robust segmentation performance.}}
\centering
{
\renewcommand\arraystretch{1.25}
\setlength{\tabcolsep}{2 mm}{
\begin{tabular}{lccccccc}\toprule[0.8pt]
     \multirow{2}{*}{Method} & Training & \multicolumn{2}{c}{{LQ3}} & \multicolumn{2}{c}{{LQ2}} & \multicolumn{2}{c}{{LQ1}} \\
                              & Time & IoU & Dice  & IoU & Dice  & IoU & Dice 
                              \\
\toprule[0.5pt]
\multirow{2}{*}{Single-stage Training} & 11h &0.617	&0.734	&0.639	&0.751	&0.648	&0.758        \\
 &  24h & 0.635	&0.749	&0.657	&0.765	&0.664	&0.770 \\
Two-stage Training (Ours) &  16h & \textbf{0.658} &   \textbf{0.765}	&\textbf{0.700}	&\textbf{0.797}	&\textbf{0.719}	&\textbf{0.812}      \\ 
\toprule[0.8pt]
\end{tabular}}
}
\label{tab:ablation-twostage-training}
\end{table*}

\begin{figure*}[t]
  \centering
   \includegraphics[width=1\linewidth]{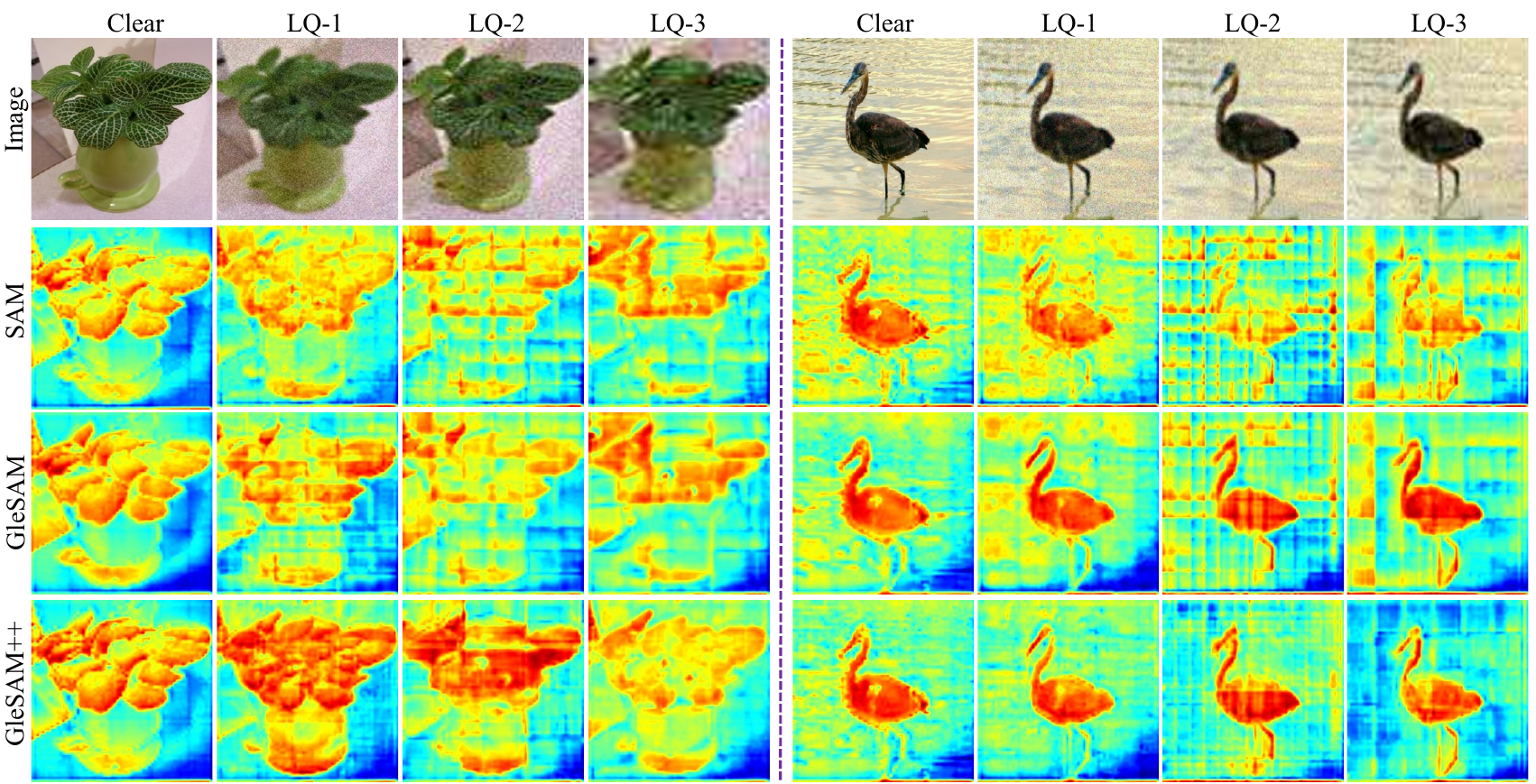}
   \caption{Visualization of feature representations for clear and degraded images under different quality levels (LQ-1, LQ-2, and LQ-3). Compared with SAM, which exhibits clear distortions and noise artifacts as degradation increases, GleSAM and GleSAM++ maintain more stable and semantically consistent feature distributions. Notably, GleSAM++ preserves object structures even under severe degradations, highlighting its strong robustness and generative enhancement capability.}
   \label{fig:sm-lqfeat}
   \vspace{-10pt}
\end{figure*}

\begin{figure*}[t]
  \centering
   \includegraphics[width=1\linewidth]{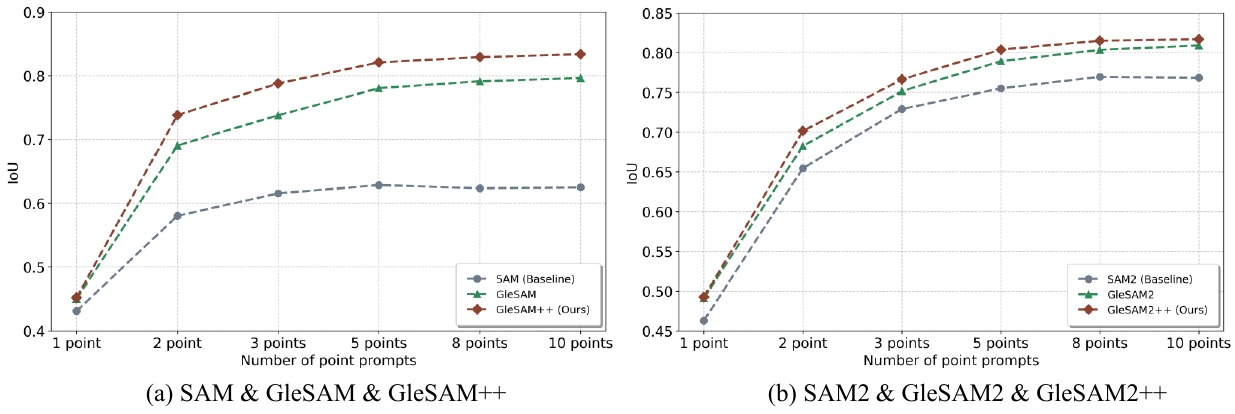}
   \caption{{Performance comparison of interactive segmentation with varying quantities of input points on the ECSSD dataset. GleSAMs consistently outperform SAMs across a range of point counts, demonstrating a more significant improvement.}}
   \label{fig:sm-inter-seg}
   \vspace{-5pt}
\end{figure*}

\begin{table}[t] \footnotesize
\caption{{Performance comparison under different prompts. We use GT-Box and Noise-Box as prompts.  The GT-Box is generated based on the GT-mask, while the Noise-Box is obtained by adding noise to the GT-Box with the noise-scale of 0.2, following \cite{dndetr}.  We present results on three different quality-degraded datasets from ECSSD, demonstrating the robustness of our method under various prompts.}}
\centering
\renewcommand\arraystretch{1.25}
\setlength{\tabcolsep}{0.9 mm}{
\begin{tabular}{ccccccc|cccccc}\toprule[0.8pt]
     \multirow{3}{*}{{Method}} & \multicolumn{6}{c|}{{\textbf{\textit{GT-Box}}}} & \multicolumn{6}{c}{{\textbf{\textit{Noise-box}}}} \\ 
     & \multicolumn{2}{c}{{LQ-3}} & \multicolumn{2}{c}{{LQ-2}} & \multicolumn{2}{c|}{{LQ-1}} & \multicolumn{2}{c}{{LQ-3}} & \multicolumn{2}{c}{{LQ-2}}  & \multicolumn{2}{c}{{LQ-1}} \\ 
     \cmidrule(r){2-3}  \cmidrule(r){4-5} \cmidrule(r){6-7}  \cmidrule(r){8-9}  \cmidrule(r){10-11} \cmidrule(r){12-13}
                               & IoU  & Dice & IoU & Dice & IoU  & Dice & IoU & Dice & IoU  & Dice & IoU & Dice \\
\toprule[0.5pt]
SAM   &0.707 & 0.812 & 0.763 & 0.852 & 0.804 & 0.881 & 0.626 & 0.744 & 0.661 & 0.770 & 0.688 & 0.788 \\
{GleSAM} & 0.754  & \textbf{0.847} &  0.804 & 0.881 & 0.839 & 0.905 & 0.686 & 0.789 & 0.732 & 0.825 & 0.763 & 0.847 \\
{GleSAM++} & \textbf{0.762} & 0.840 & \textbf{0.822} & \textbf{0.887} &  \textbf{0.857} & \textbf{0.915} & \textbf{0.698} & \textbf{0.790} & \textbf{0.745} & \textbf{0.826} & \textbf{0.775} & \textbf{0.848} \\
\midrule[0.5pt]
SAM2 &0.785 & 0.868 & 0.836 &0.902 & 0.864 & 0.921 & 0.698 &0.801 & 0.744 & 0.834 &0.759 &0.842 \\
{GleSAM2} & {0.788} & {0.870} & {0.840} & {0.905} & {0.866} & {0.922} & {0.688}  & {0.793}& {0.723}  & {0.816}  & 0.742  & 0.828  \\      
{GleSAM2++} & \textbf{0.801} & \textbf{0.883} & \textbf{0.857} & \textbf{0.924} & \textbf{0.880} & \textbf{0.939} & \textbf{0.698} & \textbf{0.801} & \textbf{0.747} & \textbf{0.836} & \textbf{0.764} & \textbf{0.848} \\        
\toprule[0.8pt]
\end{tabular}}
\label{tab:sm-box-prompt}
\vspace{-15pt}
\end{table}

\section{Further Experiments}
\label{sec: analysis}

\subsection{Visualization of feature representation}
To evaluate the capability of our method in reconstructing high-quality representations from degraded inputs, we visualize and compare the feature maps for both clear and low-quality images, as illustrated in Fig.~\ref{fig:sm-lqfeat}. 
The results show that SAM’s features are highly sensitive to degradations, exhibiting significant distortion as image quality decreases.
GleSAM alleviates this issue but remains unstable, occasionally losing fine-grained structural details under severe degradation.
In contrast, the features reconstructed by our method closely resemble the clear features, effectively recovering structural and semantic details. This demonstrates the effectiveness of our approach in enhancing feature representations and ensuring robustness in degraded scenarios.

\subsection{Comparison of varying number of point prompts} Fig.~\ref{fig:sm-inter-seg} illustrates the interactive segmentation performance under different numbers of point prompts on the ECSSD dataset.
GleSAM++ and GleSAM2++ consistently outperform SAM and SAM2 across different numbers of point prompts (from 1 point to 10 points). Note that as the prompt contains less ambiguity (with more input points), the relative performance improvement becomes more significant. This indicates GleSAM++'s robust segmentation capability.

\subsection{Comparison of other prompts}
In addition to point-based prompts, we conducted a comprehensive evaluation of our method using alternative prompting strategies, including GT-Box and Noise-Box. 
The GT-Box is directly derived from the ground truth mask, while the Noise-Box is generated by perturbing the GT-Box with a noise scale of 0.2 \cite{dndetr}, simulating scenarios with imperfect or noisy input. 
As shown in Tab. \ref{tab:sm-box-prompt}, both GleSAM++ and GleSAM2++ exhibit consistent superiority over their respective baselines across all image quality levels and prompt types. This robustness stems from the enhanced latent space representations, which mitigate noise-induced ambiguities during segmentation. Such adaptability suggests that the proposed approach generalizes well to diverse interaction scenarios and remains resilient to prompt uncertainty.

\begin{table}[t] \small
\caption{Training and inference comparison among our GleSAM++, RobsutSAM, and SAM. GleSAM++ achieves a favorable balance between performance and efficiency compared with SAM and RobustSAM.}
\centering
\renewcommand\arraystretch{1.2}
\setlength{\tabcolsep}{2.4 mm}{
\begin{tabular}{ccccccc}\toprule[0.8pt]
     \multirow{2}{*}{{Method}}  & Learnable   & Num. &  Training & Inference & Average\\
    & {Parameters} & GPUs  &  Time (h) & Speed (s) & IoU \\
\toprule[0.5pt]
SAM & -  & 256 & N/A  & 0.32 & 0.540\\
RobustSAM & 403 M & 8 & 30 h & 0.36 (+0.04) & 0.605 \\
{GleSAM++ (Ours)} & 47 M   & 4 & 16 h & 0.38 (+0.06) & \textbf{0.691} \\
\toprule[0.8pt]
\end{tabular}}
\label{tab:speed}
\vspace{-15pt}
\end{table}

\begin{figure*}[t]
  \centering
   \includegraphics[width=1\linewidth]{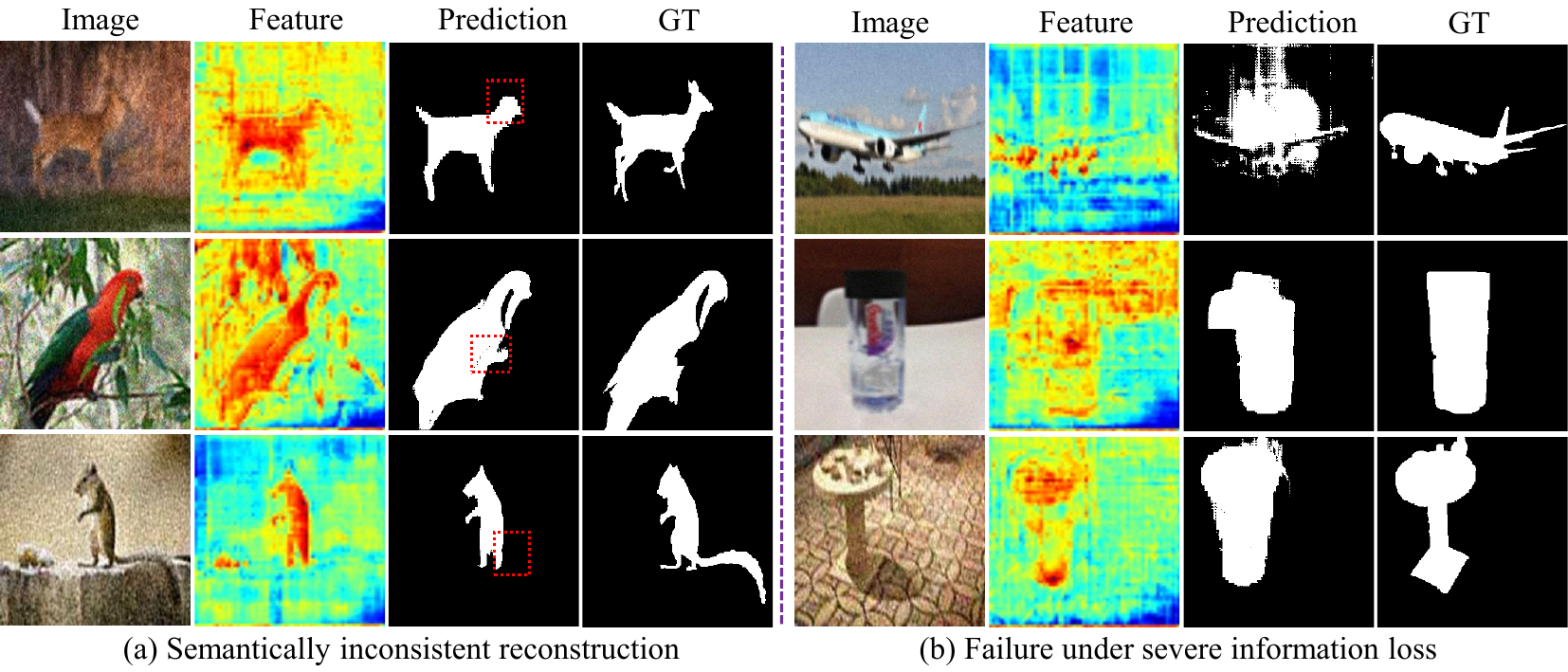}
   \caption{{Representative failure cases of GleSAM++ under severe degradations. The main errors fall into two categories: semantically incorrect but visually plausible reconstruction, and failure caused by extreme information loss.}}
   \label{fig:faliure-case}
\end{figure*}

\subsection{Analysis of computational requirements}
{ 
GleSAM++ achieves a favorable trade-off between robustness and efficiency compared with SAM and RobustSAM. Although our framework incorporates a pre-trained diffusion U-Net, the computational overhead remains limited because the generative backbone is used mainly as a source of pre-trained priors and is adapted in a parameter-efficient manner, rather than being fully fine-tuned. Moreover, our method performs enhancement directly in the latent space and adopts single-step denoising at inference time, which avoids both the heavy cost of pixel-space restoration and the iterative sampling overhead of conventional diffusion models. Consequently, as shown in Tab. \ref{tab:speed}, GleSAM++ requires only 47M learnable parameters and can be trained within 16 hours on 4 GPUs, while adding only 0.06 s latency over SAM during inference. These results demonstrate that the introduced generative module remains computationally practical and does not substantially increase the overall system complexity.
}

{
\subsection{Faliure case analysis}
\label{sec:faliure-analysis}
Fig.~\ref{fig:faliure-case} shows representative failure cases of GleSAM++ under severe degradations. We observe two major failure modes. The first is \emph{visually plausible but semantically incorrect reconstruction}. In such cases, the enhanced latent representation appears structurally reasonable, but due to insufficient semantic guidance under heavy degradation, the restored features deviate from the true object semantics, resulting in inaccurate object shapes or incomplete segmentation regions. The second is \emph{failure caused by extreme information loss}. When the degraded input loses too much structural and boundary information, the enhancement module can no longer reliably recover the correct target representation, leading to obvious reconstruction and segmentation errors.}

\section{Discussion}
\subsection{Practical Implications}
The proposed GleSAM++ enhances the robustness of segmentation under severe image degradation, which has broad practical implications across multiple vision domains.
For \textbf{high-level vision applications}, such as autonomous driving and robotic perception, accurate segmentation under adverse conditions (\eg, motion blur, or weather degradation) provides reliable scene understanding and safety-critical decision support. GleSAM++ offers an effective solution, enabling perception systems to maintain stable performance in real-world environments where image quality varies dramatically.
For \textbf{low-level vision tasks}, the reconstructed high-quality latent representations can serve as perceptual priors for enhancement and restoration tasks. By injecting structure-aware segmentation cues, GleSAM++ facilitates better spatial consistency and content fidelity in generative restoration pipelines.

\subsection{Future and Prospects} 
The proposed GleSAM++ opens several promising directions for future exploration: {(1) the generative latent-space enhancement paradigm can be further extended to other vision tasks beyond image segmentation, such as video object segmentation with SAM2 \cite{sam2}. Although our framework is structurally compatible with SAM2, the current study focuses on robustness at the frame-level representation stage under degraded-image settings, rather than in a full video segmentation scenario. Specifically, the latent enhancement is applied before mask decoding and does not explicitly modify SAM2’s temporal memory propagation mechanism. Therefore, the current results mainly verify backbone compatibility, while the effect of latent enhancement on temporal consistency across consecutive degraded frames remains to be studied. A rigorous investigation of this issue would require a dedicated degraded-video benchmark and suitable temporal evaluation metrics.}
(2) It can also be generalized to other foundation models (such as DINO series \cite{DINO-representation, dinov2, dinov3}) to improve their robustness and adaptability under challenging conditions. 
(3) Our approach can further be explored to implemented into multimodal models, such as vision-language foundation models (VLM) \cite{clip} or multi-modal large language models (MLLM) \cite{qwen2, internvl}, holding great potential for developing degradation-aware multimodal agents capable of both perception and reasoning under real-world degradations.
We hope this work inspires further research on enhancing model robustness, ultimately contributing to the reliable deployment of vision systems in real-world scenarios.

\subsection{{Limitations}}
Despite the strong experimental performance of GleSAM++, our method still has several limitations that warrant further investigation.
{
First, the generative latent-space enhancement paradigm carries an inherent risk of feature hallucination. 
In cases of extreme degradation where structural information is irretrievably lost, the latent diffusion model (LDM) may synthesize visually plausible but semantically inconsistent representations to compensate for missing details (as analyzed in Sec. \ref{sec:faliure-analysis}). Such behavior can result in confident yet incorrect segmentations, which pose potential risks in safety scenarios that demand high-fidelity reconstruction.}
Second, while the used one-step generative enhancement significantly mitigates the computational overhead typical of diffusion-based frameworks (with only 0.06 s additional latency over SAM), it still leaves room for further optimization. 
Future research could explore more lightweight generative mechanisms or adaptive inference strategies to reduce latency and memory consumption without sacrificing reconstruction fidelity.

\section{{Conclusion}}
We present GleSAM++, a framework designed to enhance SAMs for robust segmentation across images of arbitrary quality. 
By integrating the generative capability of pre-trained diffusion models into the latent space of SAMs, GleSAM++ restores degraded features and promotes more resilient segmentation performance. 
To ensure compatibility, we introduce Latent Space Alignment, including Feature Distribution Alignment for latent feature harmonization and Channel Replication and Expansion for structural consistency. 
We further introduce Degradation-aware Adaptive Enhancement to improve the adaptability of the original framework.
Finally, we construct the LQSeg dataset, which encompasses diverse degradation types and levels, enabling comprehensive training and evaluation. Extensive experiments demonstrate that GleSAM++ not only achieves superior performance on degraded images while preserving generalization to clear ones but also exhibits strong adaptability to unseen degradations. 
We believe GleSAM++ offers a promising direction toward building generation-based degradation-robust foundational segmentation models.

\clearpage

\section*{Declarations}

\noindent\textbf{Conflict of interest statement:} The authors have no relevant financial or non-financial interests to disclose.  

\noindent\textbf{Funding information:} The authors did not receive support from any organization for the submitted work.  

\noindent\textbf{Data Availability:}  The LQSeg dataset constructed in this study is publicly available at HuggingFace: \url{https://huggingface.co/guogq/GleSAM}.

\noindent\textbf{Compliance with Ethical Standards:} This article does not contain any studies with human participants or animals performed by any of the authors.


\bibliography{sn-bibliography.bib}

\end{document}